\def\year{2022}\relax
\documentclass[letterpaper]{article} 
\usepackage{aaai22}  
\usepackage{times}  
\usepackage{helvet}  
\usepackage{courier}  
\usepackage[hyphens]{url}  
\usepackage{graphicx} 

\usepackage{natbib}  
\usepackage{caption} 
\usepackage{amsmath}

\usepackage{amssymb}
\usepackage{color} 
\usepackage{colortbl}
\usepackage{hyperref}
\usepackage{multirow}
\usepackage{caption}
\usepackage{subcaption}
\usepackage{tikz}
\usepackage{comment}
\usepackage{url}
\usepackage{booktabs}
\usepackage{bbold}

\renewcommand{\eqref}[1]{(\ref{#1})}

\definecolor{Gray}{gray}{0.9}

\usepackage{xcolor}
\definecolor{mydarkblue}{rgb}{0,0.08,0.45}
\definecolor{myblue}{RGB}{103,169,207}
\definecolor{myred}{RGB}{239,138,98}
\usepackage{hyperref}
\hypersetup{%
colorlinks=true,
linkcolor=mydarkblue,
citecolor=mydarkblue,
filecolor=mydarkblue,
urlcolor=mydarkblue}

\usepackage{blindtext}

\usepackage{algorithm}
\usepackage{listings}
\usepackage{etoolbox}
\makeatletter
\AfterEndEnvironment{algorithm}{\let\@algcomment\relax}
\AtEndEnvironment{algorithm}{\kern2pt\hrule\relax\vskip3pt\@algcomment}
\let\@algcomment\relax
\newcommand\algcomment[1]{\def\@algcomment{\footnotesize#1}}
\renewcommand\fs@ruled{\def\@fs@cfont{\bfseries}\let\@fs@capt\floatc@ruled
  \def\@fs@pre{\hrule height.8pt depth0pt \kern2pt}%
  \def\@fs@post{}%
  \def\@fs@mid{\kern2pt\hrule\kern2pt}%
  \let\@fs@iftopcapt\iftrue}
\makeatother

\usepackage{amsmath,bm}
\usepackage{threeparttable}

\usepackage{newtxmath}

\DeclareCaptionStyle{ruled}{labelfont=normalfont,labelsep=colon,strut=off} 
\usepackage{algorithm}
\usepackage{algorithmic}

\usepackage{newfloat}
\usepackage{listings}
\floatstyle{ruled}
\newfloat{listing}{tb}{lst}{}
\floatname{listing}{Listing}
\title{Self-Supervised Robust Scene Flow Estimation via the Alignment of \\ Probability Density Functions}
\author{
    Pan He, Patrick Emami, Sanjay Ranka, Anand Rangarajan
}
\affiliations{
    Department of Computer and Information Science and Engineering, University of Florida
\\

    432 Newell Dr, Gainesville, FL 32611, USA\\
    pan.he,pemami@ufl.edu;  ranka,anand@cise.ufl.edu
}

\usepackage{bibentry}

\begin{document}

\maketitle

\begin{abstract}
In this paper, we present a new self-supervised scene flow estimation approach for a pair of consecutive point clouds. 
The key idea of our approach is to represent discrete point clouds as continuous probability density functions using Gaussian mixture models. Scene flow estimation is therefore converted into the problem of recovering motion from the alignment of probability density functions, which we achieve using a closed-form expression of the classic Cauchy-Schwarz divergence. 
Unlike existing nearest-neighbor-based approaches that  use hard pairwise correspondences, our proposed approach establishes soft and implicit point correspondences between point clouds and generates more robust and accurate scene flow in the presence of missing correspondences and outliers.
Comprehensive experiments show that our method makes noticeable gains over the Chamfer Distance and the Earth Mover’s Distance in real-world environments and achieves state-of-the-art performance among self-supervised learning methods on FlyingThings3D and KITTI, even outperforming some supervised methods with ground truth annotations.
\end{abstract}

\section{Introduction}

3D scene understanding \citep{qi2017pointnet,ilg2017flownet,liu2019flownet3d,dgcnn,choy20194d,he2021learning} of a dynamic environment has drawn increasing attention recently due to its wide applications in virtual reality, robotics, and autonomous driving. 
One fundamental task is \textit{scene flow estimation} that aims at obtaining a 3D motion field of a dynamic scene \citep{vedula1999three}. 
Traditional scene flow methods focus on learning representations from stereo or RGB-D images \citep{basha2013multi,jaimez2015motion,teed2021raft}. 
Recently, researchers have started to design deep scene flow estimation networks for 3D point clouds \citep{gu2019hplflownet,liu2019flownet3d,wu2019pointpwc,puy20flot,mittal2020just,gojcic2021weakly,he2021learning}.

However, major scene flow approaches rely on supervised learning with massive labeled training data that are expensive and difficult to obtain in real-world environments \citep{gu2019hplflownet,liu2019flownet3d,puy20flot}.
Consequently, researchers have turned to model training  with synthetic data and rich annotations followed by a further fine-tuning step if necessary, or the use of self-supervised learning objectives to eliminate any dependence on labels.
Early attempts at self-supervised scene flow estimation assume that the scene flow can be approximated by a point-wise transformation that moves the source point cloud to the target one \citep{wu2019pointpwc,mittal2020just,Kittenplon_2021_CVPR}.
The alignment of point clouds is measured by popular similarity metrics such as the Chamfer Distance (CD) or Earth Mover’s Distance (EMD).
However, for scene flow estimation, these metrics are limited. CD is sensitive to outliers due to its nearest neighbor criterion and tends to obtain a degenerate solution as discussed in \citet{mittal2020just} and EMD is computationally heavy and its approximations can achieve poor performance in practice.

This paper presents a principled scene flow estimation objective that addresses both limitations; it is robust to missing correspondences and outliers \emph{and} is efficient to compute. 
We accomplish this by proposing to represent discrete point clouds as continuous probability density functions (PDFs) using Gaussian mixture models (GMMs) and recovering motion by minimizing the divergence between two GMMs. 
This is in contrast to previous nearest-neighbor-based objectives which assume the existence of  hard correspondences between pairs of discrete points. 
Intuitively, if point clouds are aligned well to each other, their resulting mixtures should be statistically similar. 
We, therefore, can obtain the approximated scene flow with a decent alignment between the source and target point clouds. 
In summary, our contributions are:

\begin{itemize}
    \item A new perspective on self-supervised scene flow estimation as minimizing the divergence between two GMMs. The obtained soft correspondence between point cloud pairs differs from the existing nearest-neighbor-based approaches with the assumption of an explicit hard correspondence.     
    \item A self-supervised objective that leverages the Cauchy-Schwarz divergence for aligning two GMMs. It admits an efficient closed-form expression and leads to more robust and accurate flow estimation over CD and EMD in the presence of missing correspondences and outliers on real-world datasets.
    \item State-of-the-art performance compared to other advanced self-supervised learning methods, even outperforming some fully-supervised models that use ground truth annotations. 
\end{itemize}

\section{Related Work}
\subsection{Distance Measures for Point Clouds}
Here we provide an overview of some representative distance measures for point clouds.

\paragraph{Chamfer Distance:} Given two point sets $\mathcal{P}_1$ and $\mathcal{P}_2$, the widely-used CD \citep{fan2017point,liu2019flownet3d} is one variant of the Hausdorff distance \citep{rockafellar2009variational}, which is defined as
\begin{equation}
\footnotesize
\begin{split}
    D_{CD}(\mathcal{P}_1, \mathcal{P}_2) = & \frac{1}{|\mathcal{P}_1|} \sum_{x \in \mathcal{P}_1} \min_{y \in \mathcal{P}_2} d^2(x,y) \\ + & \frac{1}{|\mathcal{P}_2|} \sum_{y \in \mathcal{P}_2} \min_{x \in \mathcal{P}_1} d^2(x,y).
\end{split}
\end{equation}

Due to the unconstrained nearest neighbor search in CD, more than one point in $\mathcal{P}_1$ can link to the same point in $\mathcal{P}_2$ and vice versa. This many-to-one correspondence leads to noisy training signals due to the improper matching of outlier points, which are common in sparse and noisy LiDAR point clouds collected for popular autonomous driving datasets. 
Besides, as shown in \citet{mittal2020just}, due to potentially large errors in scene flow prediction, given a source point $p$, estimated scene flow $f_{est}$, and ground truth scene flow $f_{gt}$, the nearest neighbor $\mathcal{N}(\tilde{p})$ of the translated point $\tilde{p}=p+f_{est}$ may not necessarily be the same as the real translated point $\hat{p}=p+f_{gt}$, implying that $\mathcal{N}(\tilde{p}) \not\equiv \hat{p}$ and noisy training signals indeed exist.

\paragraph{Earth Mover's Distance:} A popular and efficient approximation of EMD \citep{rubner2000earth,fan2017point} establishes a one-to-one correspondence, i.e., a bijection $\phi$ mapping $\mathcal{P}_1$ to $\mathcal{P}_2$,  such that the sum of distances between \emph{all} point pairs is minimal:
\begin{equation}
\footnotesize
    D_{EMD}(\mathcal{P}_1, \mathcal{P}_2) =  \min_{\phi: \mathcal{P}_1 \rightarrow \mathcal{P}_2 } \sum_{x \in \mathcal{P}_1} d(x,\phi(x)).
\end{equation}
Obtaining an optimal mapping in EMD is non-trivial and suffers from high computational cost forcing researchers to introduce multiple approximations \citep{bertsekas1992auction,fan2017point,atasu2019linear}.
 Besides, because the approximate EMD focuses on a global point alignment (unlike $D_{CD}$), it tends to ignore local details when obtaining the optimal mapping---wrongly matching some points to distant points.
 Beyond these, other feature-based distance measures are explored for describing point cloud distances such as PointNet \citep{qi2017pointnet} and 3DmFV \citep{ben20183dmfv}.
 
As discussed above, both the CD and EMD can be sensitive to outliers and can amplify model estimation errors in a negative feedback loop during training. 
We address the limitations of these similarity measures by presenting a new differentiable objective function for scene flow estimation in the presence of noise and outliers.

\subsection{End-to-End Scene Flow Estimation}

Directly estimating scene flow from point cloud pairs using deep learning architectures has become a fast growing research area.

\noindent \textbf{Fully-Supervised Approaches}. 
FlowNet3D \citep{liu2019flownet3d} follows the PointNet++ architecture \citep{qi2017pointnet++} and introduces the \textit{flow embedding} layer to aggregate  spatio-temporal relationships of point sets based on feature similarities.  
HPLFlowNet \citep{gu2019hplflownet} projects points onto permutohedral lattices and conducts bilateral convolutions for efficient processing, leading to competitive results. 
FLOT \citep{puy20flot} follows optimal transport \cite{peyre2019computational} to establish soft point correspondences and further refine scene flow via a dynamic graph. 
FlowStep3D \citep{Kittenplon_2021_CVPR} and PV-RAFT \citep{wei2021pv}  iteratively refine scene flow predictions following \citep{teed2020raft}.

\noindent \textbf{Self-Supervised Approaches}.
In \citep{mittal2020just}, they present a self-supervised scene flow approach that addresses some limitations of CD using cycle consistency, which is orthogonal to our objective. 
PointPWC-Net \citep{wu2019pointpwc} takes inspiration from the \textit{cost volume} for optical flow estimation \cite{sun2018pwc} and generalizes the concept in a coarse-to-fine scene flow architecture. 
In \citep{pontes2020scene}, the authors regularize scene flows from point clouds via graph construction and application of the graph Laplacian. 
An  adversarial learning approach for scene flow estimation \citep{zuanazzi2020adversarial} maps point clouds to a latent space in which a robust distance metric can be computed. 
Self-Point-Flow \citep{li2021self} effectively leverages optimal transport \citep{peyre2019computational} to generate noisy pseudo ground truth and refines it via a random walk, in contrast to our end-to-end self-supervised objective with a closed-form expression.

Most existing self-supervised scene flow approaches \citep{wu2019pointpwc,mittal2020just,Kittenplon_2021_CVPR} explicitly establish hard point correspondences between discrete point clouds based on the nearest neighbor criterion. 
Unlike them, we represent discrete point clouds as continuous PDFs, i.e., GMMs, and establish an implicit soft point correspondence via minimization of the divergence between the two corresponding mixtures.
Our approach is less sensitive to the missing correspondences (e.g., due to occlusion or view changes) and outliers.

\subsection{Point Set Registration}
 
Kernel correlation-based registration was proposed in \citep{tsin2004correlation} to align two point sets seen as PDFs. 
The method is further improved and generalized in \citep{roy2007deformable,jian2010robust,hasanbelliu2011robust} with other divergences. 
Recently, deep learning methods have obtained impressive results for point set registration
\citep{wang2019deep,aoki2019pointnetlk,yew2020rpm}. 

DeepGMR \citep{yuan2020deepgmr} presents a deep registration method that minimizes the Kullback–Leibler divergence \citep{kullback1951information} between point clouds.
Our approach is related to these as we also represent discrete point clouds as PDFs to align them.

\section{Problem Definition}
Point clouds can represent raw data, e.g., 3D shapes, or the surfaces from which they are sampled, e.g., those collected or reconstructed from LiDAR or RGB-D sensors. 
Our goal is to estimate 3D scene flow from consecutive point cloud frames.
Denote the source point cloud as $\boldsymbol{S}= \{ (\boldsymbol{c_i^s}, \boldsymbol{x_i^s}) \mid i=1,\dots, N\}$ and target point cloud as $\boldsymbol{T}= \{ (\boldsymbol{c_j^t}, \boldsymbol{x_j^t}) \mid j=1,\dots, M\}$, where $\boldsymbol{c_i^s}, \boldsymbol{c_j^t}$ are the 3D coordinates of individual points and  $\boldsymbol{x_i^s}, \boldsymbol{x_j^t}$ are the associated point features, e.g., color or LiDAR intensity. 
Due to the viewpoint shift, occlusion and sampling effect, $\boldsymbol{S}$ and $\boldsymbol{T}$ do not necessarily have the same number of points or have strict point-to-point correspondences. 
Considering points $\boldsymbol{s_i} = (\boldsymbol{c_i^s}, \boldsymbol{x_i^s})$ in the source point cloud $\boldsymbol{S}$ being moved to a new location $\boldsymbol{\widehat{c_i^s}}$ at the target frame and denoting its 3D motion as $\boldsymbol{d_i} = \boldsymbol{\widehat{c_i^s}} - \boldsymbol{c_i^s}$, a scene flow estimation model will predict the motion for every sampled point $\boldsymbol{s_i}$ in the source point cloud $\boldsymbol{S}$ via a function $f$:  $\boldsymbol{D} = \{ \boldsymbol{d_i} = f(\boldsymbol{S}, \boldsymbol{T})_i \mid i=1,\dots, N\}$ such that they are close to real motion.

\section{Proposed Approach}

In this section, we introduce the proposed approach for representing and aligning discrete point clouds using PDFs. 
To the best of our knowledge, this is the first attempt at doing so for scene flow estimation. 

\subsection{Mixture Models for Representing Point Clouds}
Unlike existing self-supervised learning objectives such as CD and EMD that rely on hard pairwise correspondences between discrete point clouds, the key idea of our paper is to represent point clouds by PDFs to obtain a soft correspondence. 
We demonstrate the conceptual differences between CD, EMD, and CS in Figure~\ref{fig:cd_emd_cs_curve}. 
The main intuition is that point clouds can be interpreted as samples drawn from continuous spatial distributions of point locations. 
By doing so, we can capture the uncertainty in point cloud generation, e.g., jitter introduced during the LiDAR scanning process. 

Formally, for a given point cloud $\boldsymbol{x}$, we represent it as the PDF of a general Gaussian mixture, which is defined as  $\mathcal{G}(x) = \sum_{k=1}^K w_k \mathcal{N}(x | \boldsymbol{\mu_k}, \boldsymbol{\Sigma_k})$ with
\begin{equation}
    \mathcal{N}(x| \boldsymbol{\mu_k}, \boldsymbol{\Sigma_k}) = \frac{\exp\left[-\frac{1}{2}(x-\boldsymbol{\mu_k})^T \boldsymbol{\Sigma_k}^{-1}(x-\boldsymbol{\mu_k})\right]}{\sqrt{ (2\pi)^d |\boldsymbol{\Sigma_k}|}},
\end{equation}
where $K$ is the number of Gaussian components. We denote $w_k, \mu_{k}, \Sigma_{k}$  as the mixture coefficient, mean, and covariance matrix  of the $k^{th}$ component of $\mathcal{G}(x)$. 
$d$ is the feature dimension of each point. 
In our case, $d=3$. $|\boldsymbol{\Sigma_k}|\equiv \det \boldsymbol{\Sigma_k}$ is the determinant of $\boldsymbol{\Sigma_k}$, also known as the generalized variance. 
Note that if $K$ is large enough, $\mathcal{G}(x)$ can well approximate almost any underlying density of a point cloud.

Inspired by \citep{jian2010robust,roy2007deformable}, we simplify the GMMs as follows: 1) the number of Gaussian components is the number of points with uniform weights (the occupancy probabilities or the mixture coefficients), 2) the mean vector of a component is the location of each point, and 3) all components share the same variance (isotropic, or spherical covariances), i.e., $\Sigma_i=\Sigma_j= \sigma \boldsymbol{I}$ with the identity matrix $\boldsymbol{I}$. 
 We, therefore, obtain an overparameterized GMM model which can be equivalently obtained from a fixed-bandwidth kernel density
estimation (KDE) with a Gaussian kernel \citep{scott2015multivariate}. 
More complicated GMMs are non-trivial and would require computationally expensive model fitting such as the Expectation-Maximization (EM) algorithm \citep{moon1996expectation}, which we do not explore in this paper and instead reserve for future work.

\subsection{Recovering Motion from the Alignment of PDFs}

The principle of PDF divergence minimization results in the specification of a self-supervised learning objective that optimizes a scene flow model such that a dissimilarity measure $D_{dsim}(\mathcal{G}(\boldsymbol{S}_{w}), \mathcal{G}(\boldsymbol{T}))$ between the GMM representations of the warped point cloud $\mathcal{G}(\boldsymbol{S}_{w}) =\mathcal{G}(\boldsymbol{S}+\boldsymbol{D})$ and the target point cloud $\mathcal{G}(\boldsymbol{T})$ is minimized. 
Recall that $\boldsymbol{D} = \{ \boldsymbol{d_i} = f(\boldsymbol{S}, \boldsymbol{T})_i \mid i=1,\dots, N\}$ where $f$ is implemented as a deep neural network. 
We can construct a suitable $D_{dsim}$ such that it is differentiable so it can guide optimization via backpropagation and gradient descent. We now describe how to achieve this goal.

We choose the Cauchy-Schwarz (CS) divergence \citep{jenssen2005optimizing,principe2010information} for measuring the similarity between the two GMM representations of point clouds $\boldsymbol{S}_{w}$ and $\boldsymbol{T}$. 
The CS divergence can be \emph{expressed in closed-form}, allowing an efficient end-to-end trainable implementation for scene flow estimation. 
We optimize $f$ by minimizing
\begin{equation}
\footnotesize
\begin{split}\label{eq:cs_divergence}
    & \mathcal{D}_{CS}( \mathcal{G}(\boldsymbol{S}_{w}),\mathcal{G}(\boldsymbol{T}))  = - \log \Big( \frac{\int \mathcal{G}(\boldsymbol{S}_{w}) \mathcal{G}(\boldsymbol{T}) dx}{\sqrt{\int \mathcal{G}^2(\boldsymbol{S}_{w}) dx \int \mathcal{G}^2(\boldsymbol{T})} dx} \Big)  \\
   & = - \log  \int  \mathcal{G}(\boldsymbol{S}_{w})   \mathcal{G}(\boldsymbol{T}) dx +  0.5 \log  \int \mathcal{G}^2(\boldsymbol{S}_{w}) dx \\ &  \quad +  0.5 \log   \int  \mathcal{G}^2(\boldsymbol{T}) dx.
\end{split}
\end{equation}
The CS divergence is derived from the CS inequality \citep{steele2004cauchy} and is expressed as inner products of PDFs. 
It defines a symmetric measure for any two PDFs $\mathcal{G}(\boldsymbol{S}_{w})$ and $\mathcal{G}(\boldsymbol{T})$ such that $0 \le D_{CS} < \infty$ where the minimum is obtained iff $\mathcal{G}(\boldsymbol{S}_{w}) = \mathcal{G}(\boldsymbol{T})$.
It measures the interaction of the generated field of one PDF on the locations of the other PDF, which is also called the \textit{cross information
potential} of the two densities \citep{hasanbelliu2011robust}.
\begin{figure*}[hbt!]
    \centering
     \includegraphics[width=0.99\textwidth]{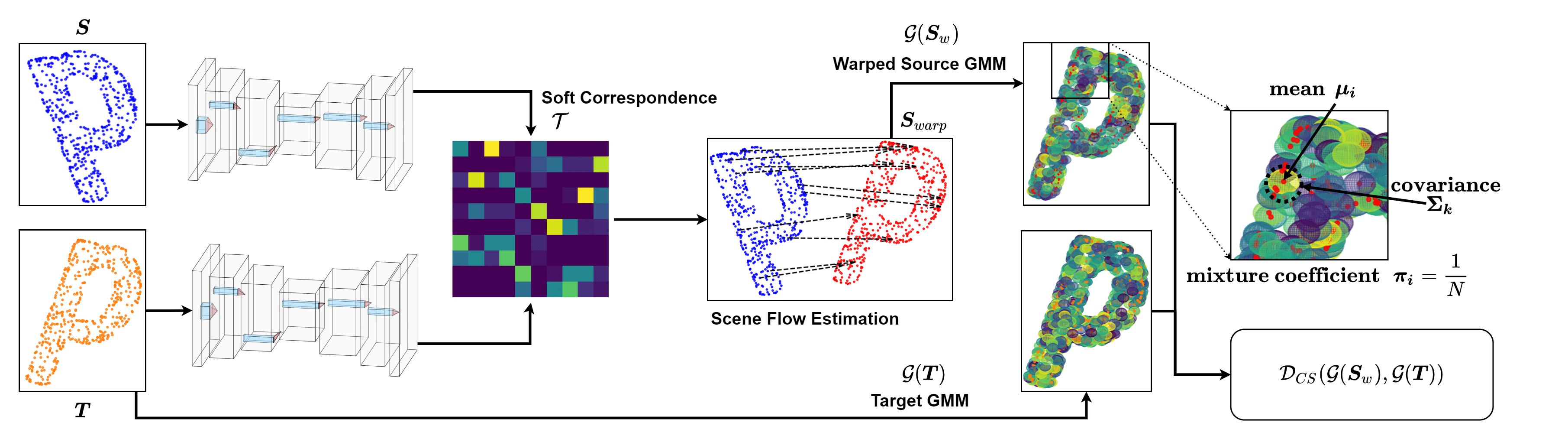}
    \caption{Overview of the proposed self-supervised learning for scene flow estimation. Our model takes both source and target point clouds to extract deep features via a UNet-like encoder-decoder backbone network  \citep{ronneberger2015u} based on MinkowskiNet \citep{choy20194d}. We then warp the source point cloud by adding the estimated scene flow. Both the warped source and target point clouds are further fit using two separate GMMs. We train the model by minimizing the discrepancy between the two corresponding mixtures via a closed-form expression for the CS divergence.}
    \label{fig:main}
\end{figure*}

\paragraph{Closed-form expression for GMMs:}
The CS divergence in Equation~\ref{eq:cs_divergence} can be written in a closed-form expression for GMMs \citep{jenssen2006cauchy}. 
The basic idea is to follow the Gaussian identity \citep{petersen2008matrix} to obtain the product of two Gaussian PDFs as
\begin{equation}
\begin{split}\label{eq:main_product}
    \mathcal{N}(\boldsymbol{x}| \boldsymbol{\mu_i}, \boldsymbol{\Sigma_i}) & \mathcal{N}(\boldsymbol{x}| \boldsymbol{\mu_j}, \boldsymbol{\Gamma_j})  = \\ & \mathcal{N}(\boldsymbol{\mu_i}| \boldsymbol{\mu_j}, \boldsymbol{\Sigma_i}  + \boldsymbol{\Gamma_j}) \mathcal{N}(x| \boldsymbol{\mu_{ij}}, \boldsymbol{\Sigma_{ij}})
\end{split}
\end{equation}
 where
 \begin{equation}
 \footnotesize
     \boldsymbol{\mu_{ij}}  = \boldsymbol{\Sigma_{ij}} (\boldsymbol{\Sigma_i}^{-1} \boldsymbol{\mu_{i}}  + \boldsymbol{\Gamma_j}^{-1} \boldsymbol{\mu_{j}})
 \end{equation}
and
\begin{equation}
 \footnotesize
\boldsymbol{\Sigma_{ij}} = {(\boldsymbol{\Sigma_i}^{-1}  + \boldsymbol{\Gamma_j}^{-1})}^{-1}.
\end{equation} 
Then we can use the Gaussian identity trick to simplify each term in the right of Equation~\ref{eq:cs_divergence} and get
\begin{equation}
\footnotesize
\begin{split}
 & \mathcal{D}_{CS}( \mathcal{G}(\boldsymbol{S}_{w}),\mathcal{G}(\boldsymbol{T})) =  - \log \bigg(\sum_{i,j=1}^{N,M}  \pi_{i}  \tau_{j}  \mathcal{N} (\boldsymbol{c^s_i} | \boldsymbol{c^t_j}, \boldsymbol{\Sigma_i} + \boldsymbol{\Gamma_j} )\bigg)  \\
         & + 0.5 \log \bigg(\sum_{i,i'=1}^{N,N}   \pi_{i}  \pi_{i'}  \mathcal{N} (\boldsymbol{c^s_i} | \boldsymbol{c^s_{i'}}, \boldsymbol{\Sigma_i} + \boldsymbol{\Sigma_{i'}}) \bigg) \\ 
         & + 0.5 \log \bigg(\sum_{j, j'=1}^{M,M}  \tau_{j}  \tau_{j'}  \mathcal{N} (\boldsymbol{c^t_j} | \boldsymbol{c^t_{j'}}, \boldsymbol{\Gamma_j} + \boldsymbol{\Gamma_{j'}}) \bigg), \label{eq:final}
\end{split}
\end{equation}
where we denote the sets of mixture coefficients for two GMMs $\mathcal{G}(\boldsymbol{S}_{w})$ and $\mathcal{G}(\boldsymbol{T})$ as $\{ \pi_{i} \}_{i=1}^N$ and $\{ \tau_{j} \}_{j=1}^M$ and the corresponding covariance matrix sets as $\{ \Sigma_{i} \}_{i=1}^N$ and $\{ \Gamma_{j} \}_{j=1}^M$. Note that the third term in the right of Equation~\ref{eq:final} is a constant value for a target point cloud and can be optionally removed for faster computation. The detailed derivation of $\mathcal{D}_{CS}( \mathcal{G}(\boldsymbol{S}_{w}),\mathcal{G}(\boldsymbol{T})) $ can be found in the appendix. 

\begin{figure}
    \centering
    \includegraphics[width=0.48\textwidth]{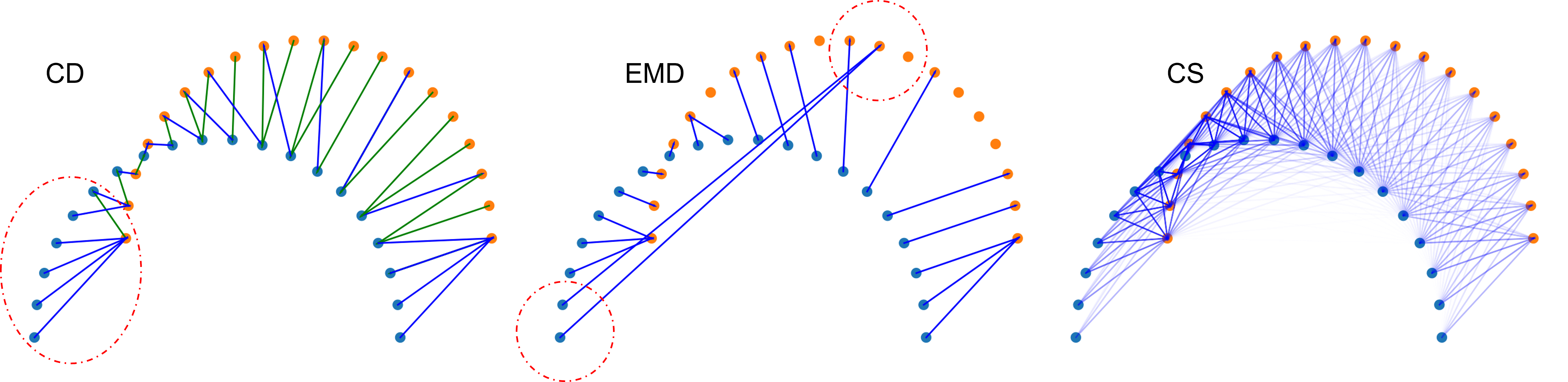}
    \caption{A toy example to illustrate the conceptual difference between CD, EMD, and CS. For CD and EMD, we visualize their matching correspondences between points of blue and orange curves by adding blue (the forward matching) and green (the backward matching) arrows. CD and the EMD approximation find the hard pairwise correspondence while \citep{fan2017point} CS tries to link every blue point to all orange points via Gaussian functions --- here we only show $\mathcal{N} (\boldsymbol{c^s_i} | \boldsymbol{c^t_j}, \boldsymbol{\Sigma_i} + \boldsymbol{\Gamma_j} )$ in Equation \ref{eq:final} and weight the arrow color according to their values.}
    \label{fig:cd_emd_cs_curve}
\end{figure}
\paragraph{Discussion:} Examining $\mathcal{D}_{CS}( \mathcal{G}(\boldsymbol{S}_{w}),\mathcal{G}(\boldsymbol{T}))$, we can see that the exponential function in the Normal PDFs applied to the square of the point-pair distances weighed by a combination of $\Sigma_{i}$ and $\Gamma_{j}$ mitigates over-sensitivity to outliers by suppressing large squared distances. 
This is in contrast to the $\ell_2$ distance-based approaches such as CD and EMD where outliers can contribute large values to the loss and negatively impact training. 
Also, it is fully differentiable and can be easily implemented with a few lines of code, which we provide in the appendix, in contrast to the sub-differentiable CD due to the $\min$ operator and the sub-optimal EMD approximation with expensive computation \citep{fan2017point}. 
Due to the probabilistic formulation, it supports the processing of point clouds with different sizes while being tolerant to noise. The conceptual difference between CD, EMD, and CS is illustrated in Figure \ref{fig:cd_emd_cs_curve}.
We note that CS is also closely related to graph cuts and Mercer kernel theory (see \citep{jenssen2006cauchy} for a detailed discussion). 

\subsection{Model Implementation}

We now demonstrate a deep neural network for self-supervised scene flow estimation that is trained with the CS divergence (Figure \ref{fig:main}). It consists of : 1) deep feature extraction, 2) scene flow estimation from soft correspondences and 3) our CS divergence objective. 
We also add other regularization techniques to regularize the unconstrained scene flow predictions by encouraging rigid motion in local regions.

\noindent \textbf{Deep Feature Extraction:} To obtain features that encode raw point coordinates into higher dimension, we utilize a UNet-like encoder-decoder backbone network  \citep{ronneberger2015u} based on MinkowskiNet \citep{choy20194d}. It receives inputs that are voxelized from point clouds. Denote these sparse voxels as $V^{s} \in \mathbb{R}^{N^s \times 3}$ and $V^{t} \in \mathbb{R}^{N^t \times 3}$ for $\mathcal{S}$ and $\mathcal{T}$ respectively. Both $V^{s}$ and $V^{t}$ will be feed into a same backbone with shared weights to obtain their corresponding deep features $F^{s} \in \mathbb{R}^{N^s \times 64}$ and $F^{t} \in \mathbb{R}^{N^t \times 64}$.

\noindent \textbf{Scene Flow Estimation:} Similar to \citep{puy20flot,wei2021pv}, we construct a cost matrix $\mathcal{C} \in \mathbb{R}^{N^s \times N^t}$ and establish a soft-correspondence between sparse voxels. Formally, we compute a cost between every point $f^s_i \in F^{s}$ and every point $f^t_j \in F^{t}$ defined as 
\begin{equation}
    \mathcal{C}_{ij} = 1 -  \frac{\langle f^s_i,f^t_j \rangle}{{\|f^s_i \|}_2{\| f^t_j \|}_2}.
\end{equation}

Based on the cost matrix $\mathcal{C}$, we find the top-K smallest values for each sparse voxel $v^s_i \in V^{s}$ regarding $V^t$, whose corresponding index set is denoted as $\mathcal{N}^k_i$.
We then estimate the scene flow of $v^s_i$  as 
\begin{equation}
    d^v_i = \frac{\sum_{j \in \mathcal{N}^k_i} \mathcal{T}_{ij} v^t_j}{\sum_{j \in \mathcal{N}^k_i} \mathcal{T}_{ij}} - v^s_i.
\end{equation}
where  $\mathcal{T}_{ij} = \exp(-\mathcal{C}_{ij}/\epsilon)$. We empirically set $\epsilon=0.00625$.
Therefore, $d^v_i$ is the weighted distance between $v^s_i$ and its top-K  points $\{v^t_j | j \in \mathcal{N}^k_i, |\mathcal{N}^k_i| \leq N^t\}$.  We apply the inverse-distance weighted interpolation to transfer the flow for the sparse voxels to each point $s_i = \{c^s_i, x^s_i\}$ in $S$:
\begin{equation}
    d^s_i = \frac{\sum_{j: v^s_j \in \mathcal{M}(s_i)} d^v_i \|v^s_j-c^s_i \|_2^{-1} }{\sum_{j: v^s_j \in \mathcal{M}(s_i)} \|v^s_j-c^s_i\|_2^{-1}},
\end{equation}
where $\mathcal{M}(s_i)$ finds the k-nearest neighbor (KNN) voxels of the point $s_i$ based on the Euclidean distance.  

\subsection{Training}
Our model is trained end-to-end without requiring any ground-truth scene flow annotations.

\noindent \textbf{The CS divergence loss:} We use the introduced CS divergence for self-supervised learning.
\begin{equation}
    E_{cs}(\boldsymbol{D}) = \mathcal{D}_{CS}( \mathcal{G}(\boldsymbol{S}_{w}),\mathcal{G}(\boldsymbol{T})) ,
\end{equation}
where we add the predicted scene flow $\boldsymbol{D}$ to the source point cloud $\mathcal{S}$ to obtain $\boldsymbol{S}_{w}$. It aligns the warped source and target point clouds.

\noindent \textbf{The graph Laplacian loss:} As each estimated scene flow vector in $\boldsymbol{D}$ has three degrees of freedom, only applying CS is under-constrained with many possible estimations. We further regularize them by adding an extra constraint to enforce the transformation to be as rigid as possible \citep{bobenko2007discrete}, which we formulate as
\begin{equation}\label{eq:rigid}
    E_r(\boldsymbol{D}) = \sum_{\{i,j\} \in E} \|d_i - d_j\|_1, 
\end{equation}
where $E$ defines the edge set of a graph $G$ built upon the source point cloud $\mathcal{S}$. 
There exist various ways of constructing the graph: KNN graphs, fixed-Radius Nearest Neighbor graphs are possibilities.
Following \citep{pontes2020scene}, we adopt the KNN graph due to its better sparsity and connectivity
properties. 
We reformulate Equation~\ref{eq:rigid} as 
\begin{equation}
    E_r(\boldsymbol{D}) = \frac{1}{|\mathcal{S}|} \sum_{i=1}^N \frac{1}{|\mathcal{I}(s_i)|} \sum_{j \in \mathcal{I}(s_i)} \|d_i - d_j\|_1,
\end{equation}
where $\mathcal{I}(s_i)$ denotes the index set of the k-nearest neighbor points of $s_i$ in $\mathcal{S}$. We empirically set $k=50$ leaving this choice up for future exploration.

\textbf{The full objective function} is a weighted sum of the CS divergence loss and the graph Laplacian loss:
\begin{equation}
    E(\boldsymbol{D}) = E_{cs}(\boldsymbol{D}) + \lambda E_r(\boldsymbol{D})
\end{equation} where $\lambda$ is a hyperparameter to balance two terms. 
The first term minimizes the discrepancy between the mixtures representing the warped source point cloud $\mathcal{S}+\boldsymbol{D}$ and the target $\mathcal{T}$. 
The second term enforces the predicted flows of nearby points to be similar to each other.

\section{Experiments} \label{sec:experiments}

In this section, we describe the datasets and the associated evaluation metrics. 
We demonstrate the advantage of the proposed CS divergence loss over CD and EMD. 
Our proposed approach achieves state-of-the-art performance compared to self-supervised learning approaches and even surpasses some fully supervised methods.

\subsection{Datasets and Evaluation Metrics}\label{sec:dataset}

\textbf{FlyingThings3D (FT3D):} The dataset \cite{mayer2016large} is the first large-scale synthetic dataset designed for scene flow estimation where each scene contains multiple randomly-moving objects taken from the ShapeNet dataset \citep{chang2015shapenet}. 
Following \citep{liu2019flownet3d,gu2019hplflownet}, we reconstructed point clouds and their ground truth scene flows, ending up with a total of $19,640$ training examples and $3,824$ test examples. 
We selected $3,928$ examples from the training set for a hold-out validation.

\noindent \textbf{KITTI Scene Flow (KSF):} This real-world dataset \citep{menze2015joint,menze2015object} is adapted from the KITTI Scene Flow benchmark with $142$ frame pairs. 
We obtained point clouds and ground truth scene flow by lifting the ground-truth disparity maps and optical flow to 3D \citep{gu2019hplflownet}. 
We removed ground points by heuristically setting a height threshold \citep{liu2019flownet3d,gu2019hplflownet}.

\noindent \textbf{Unlabelled KITTI$_{r}$ Dataset:} The KITTI$_{r}$ dataset is prepared by \citep{li2021self} where they use raw point clouds in KITTI to produce a training dataset. 
It collects samples from those scenes not included in KSF and guarantees separate training and testing splits. 
It results in $6,068$ training point cloud pairs sampled at every five frames.

\noindent \textbf{Evaluation Metrics:} We use the standard evaluation metrics \citep{liu2019flownet3d,gu2019hplflownet,puy20flot}: 1) the \textit{EPE3D[m]}, or the end-point error, which calculates the mean
absolute distance error in meters, 2) the \textit{strict ACC3D (Acc3DS)} considering the percentage of points whose $EPE3D[m]<0.05m$ or relative error $<5\%$, 3) the \textit{relaxed ACC3D  (Acc3DR)} considering the percentage of points whose $EPE3D[m]<0.1m$ or relative error $<10\%$, and 4) the \textit{Outliers3D} to compute the percentage of points whose $EPE3D[m]>0.3m$ or relative error $>10\%$.

\subsection{Implementation Details}

We randomly sampled $8,192$ points for each point cloud. All models were implemented in Pytorch \citep{paszke2019pytorch} and MinkowskiEngine \cite{choy20194d}.
We utilized the Adam optimizer \citep{kingma2014adam} with an initial learning rate of $0.001$ and the cosine annealing scheduler to gradually decrease it based on the epoch number. 
We trained models for 100 epochs and voxelized points at a resolution of 0.1m.

\subsection{Comparison with CD and EMD}
To verify the robustness of CS, we conduct one critical study of training models on KITTI$_r$. 
Specifically, to make a fair comparison, we use our model with the same architecture while applying different self-supervised objective functions including CD, EMD, and CS.  
We then evaluate these trained models on KSF with ground truth annotations. 
Because the KITTI$_r$ dataset is collected from autonomous vehicles across real-world environments with varied conditions, it tends to include more noise and outliers when compared to synthetic datasets. 
Therefore, we would expect models trained with a more robust objective function to achieve better performance. 

As shown in Table \ref{tab:robustness}, our model trained with CS outperforms both models with CD and EMD significantly. 
Specifically, it achieves a much lower EPE of $0.105$, decreasing the errors of CD and EMD by $38.24\%$ and $45.31\%$. 
It shows that CS is much more robust when conducting self-supervised learning on the real-world KITTI$_r$ dataset.
 Furthermore, it has achieved better performance compared to some representative state-of-the-art supervised methods, such as Flownet3D \citep{liu2019flownet3d}, HPLFlowNet \citep{gu2019hplflownet}, and EgoFlow \citep{tishchenko2020self}.

\subsection{Ablation Studies}

\noindent \textbf{Choices of the kernel bandwidth $\bm{\sigma}$:} Recall that the CS divergence with GMMs is closely related to fixed bandwidth KDE with Gaussian kernels.
The KDE bandwidth equals the scale parameter (standard deviation) of a Gaussian kernel. 
It is important to choose a suitable bandwidth (corresponding to the square root of the isotropic variance in GMM models)---either too large or too small bandwidth values will not best approximate the true underlying density, resulting in degraded performance.

\begin{figure}[hbt!]
    \centering
    \includegraphics[width=0.48\textwidth]{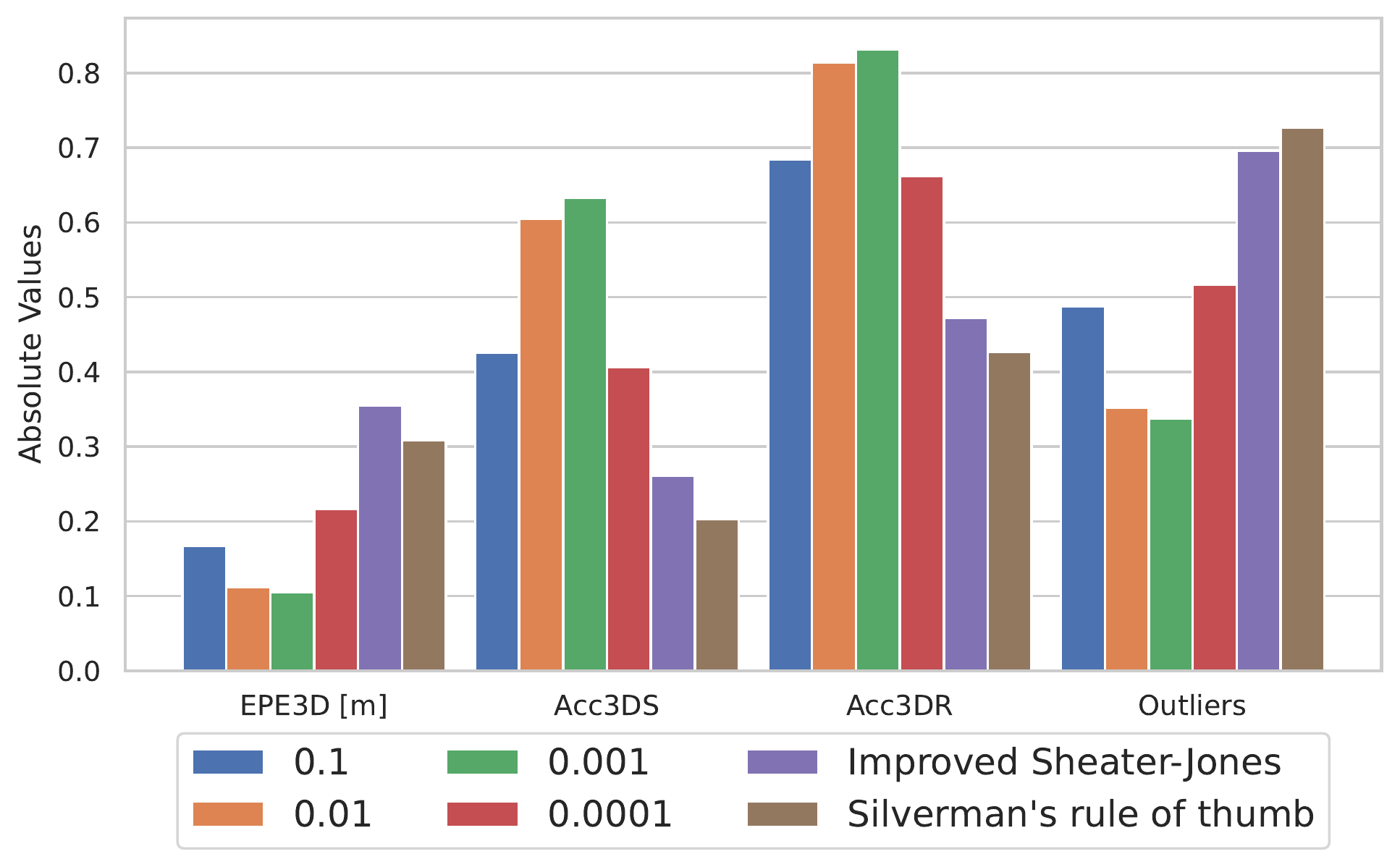}
    \caption{Results of CS with different kernel bandwidths. Choosing a suitable bandwidth leads to best performance.}
    \label{fig:bandwidth}
\end{figure}

As shown in Figure \ref{fig:bandwidth}, we plot the results by choosing different variances $\sigma^2 = 0.1, 0.01, 0.001, 0.0001$ where $\sigma$ is the kernel bandwidth. 
When setting a large variance such as $\sigma^2=0.1$, it leads to \textit{oversmoothing} where some important local structures, e.g., curvatures of a shape, are ignored due to a large amount of smoothing. 
A small variance such as $\sigma^2=0.0001$ tends to generate darting structures on a density curve which is sensitive to noise---considering its extreme situation ($\sigma \to 0$) that changes a Gaussian PDF to a Dirac delta function. 
The results show that choosing $\sigma^2=0.01$ or $\sigma^2=0.001$ leads to good results.  
Nevertheless, such a bandwidth selection can vary across datasets depending on the underlying data characteristics. 
We also explored conventional bandwidth selection methods including \textit{Silverman's rule of thumb} \citep{silverman2018density} and \textit{Improved Sheater-Jones (ISJ)} \citep{botev2010kernel}. 
They achieve worse results, which might be caused by the improper data assumption, e.g., unimodal distribution in Silverman's rule, and uncertainty not being captured.

\begin{table}
	\centering
	\caption{Comparisons between models trained with different self-supervised objective functions, i.e., CD, EMD, and CS. We evaluate all trained models on the KITTI Scene Flow dataset. \textbf{We find that CS is more robust to noise in LiDAR datasets than CD and EMD}.}\label{tab:robustness}
	\Huge
	\resizebox{\columnwidth}{!}{%
    \begin{tabular}{l|cc|cccc}
            \toprule
			Method & Sup. & Training data &EPE3D [m]~$\downarrow$ & Acc3DS~$\uparrow$ & Acc3DR~$\uparrow$ & Outliers~$\downarrow$ \\
			\noalign{\smallskip}\hline\noalign{\smallskip}
			Flownet3D~\shortcite{liu2019flownet3d} & $\mathit{Full}$ & FT3D & 0.177 & 0.374 & 0.668 & 0.527 \\
            HPLFlowNet~\shortcite{gu2019hplflownet} & $\mathit{Full}$ & FT3D  & 0.117 & 0.478 & 0.778 & 0.410 \\
			EgoFlow~\shortcite{tishchenko2020self}  & $\mathit{Full}$ & FT3D & 0.103 & 0.488 &0.822 & 0.394 \\
			\noalign{\smallskip}\hline\noalign{\smallskip}
			CD (Ours) & $\mathit{Self}$ & KITTI$_r$& 0.170         & 0.477         & 0.697         & 0.470   \\
            EMD (Ours)   & $\mathit{Self}$ & KITTI$_r$  & 0.192  & 0.426   & 0.666  & 0.503   \\
            CS (Ours) & $\mathit{Self}$ & KITTI$_r$ & \textbf{0.105} & \textbf{0.633} &   \textbf{0.832} &  \textbf{0.338} \\
            \hline
	\end{tabular}
	}

\end{table}

\begin{table}[hbt!]
	\centering
	\caption{Impact of graph Laplacian regularization. Choosing a proper $\lambda$ leads to further improvements. }\label{tab:laplacian}
	\Huge
	\resizebox{\columnwidth}{!}{%
    \begin{tabular}{l|cc|cccc}
            \toprule
			Regularization  & Sup. & Training data &EPE3D [m]~$\downarrow$ & Acc3DS~$\uparrow$ & Acc3DR~$\uparrow$ & Outliers~$\downarrow$ \\
			\noalign{\smallskip}\hline\noalign{\smallskip}
			w/o ($\lambda$=0) & $\mathit{Self}$ & KITTI$_r$ & 0.105 & 0.633 &   0.832 &  0.338 \\
			\noalign{\smallskip}\hline\noalign{\smallskip}
			$\lambda$ = 100 & $\mathit{Self}$ & KITTI$_r$& 0.109       & 0.639        & 0.824        & 0.332 \\
            $\lambda$ = 10 & $\mathit{Self}$ & KITTI$_r$  & \textbf{0.096}  & \textbf{0.686}  & \textbf{0.855}  & \textbf{0.302}   \\
            $\lambda$ = 1 & $\mathit{Self}$ & KITTI$_r$ & 0.098 & 0.659 &   0.853 &  0.316 \\
            $\lambda$ = 0.1 & $\mathit{Self}$ & KITTI$_r$ & 0.103 & 0.631 &   0.840 &  0.331 \\
            \hline
	\end{tabular}
	}

\end{table}
\begin{figure*}[hbt!]
    \centering
    \includegraphics[width=\textwidth]{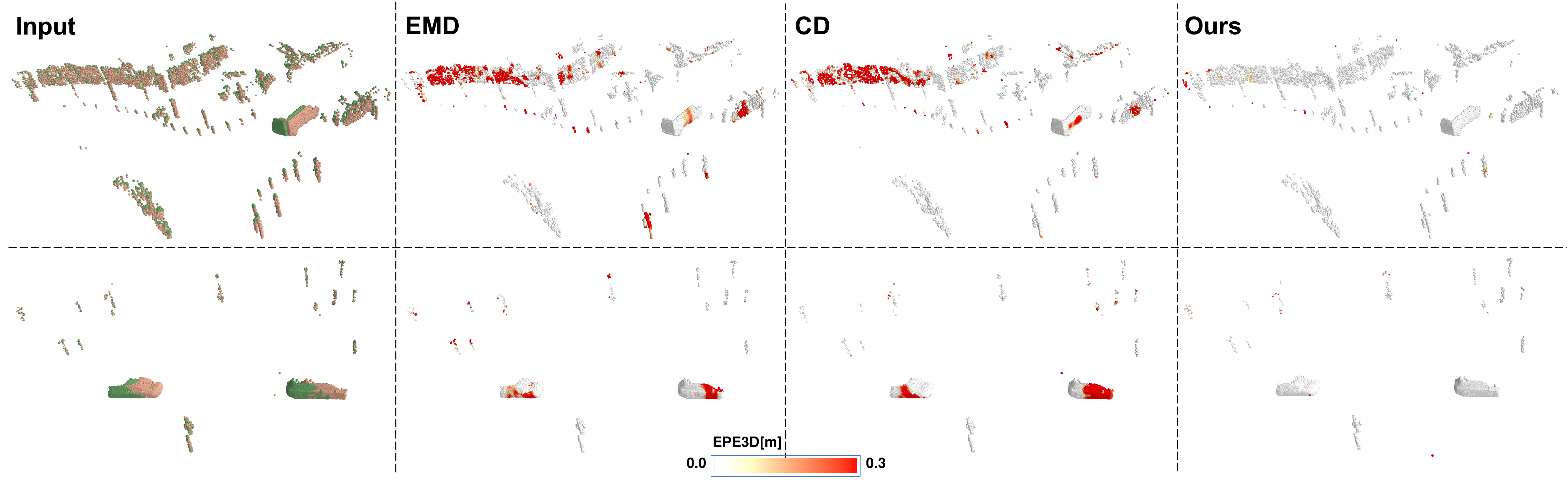}
    \caption{Qualitative results of our method on \emph{KSF}. We clip the EPE3D[m] to the range between $0.0$ m (white) and $0.3$ (red). It confirms that our method is better on handling outliers compared to EMD and CD. }
    \label{fig:ksf}
\end{figure*}
\begin{table}[hbt!]

	\centering
	\caption{Evaluation results on the \emph{FT3D} datasets.}\label{tab:ft3d}
	\Huge
	\resizebox{\columnwidth}{!}{%
    \begin{tabular}{lccccc}
            \toprule
			Method & Sup. & EPE3D[m]~$\downarrow$ & Acc3DS~$\uparrow$ & Acc3DR~$\uparrow$ & Outliers~$\downarrow$ \\
			\noalign{\smallskip}\hline\noalign{\smallskip}
             FlowNet3D~\shortcite{liu2019flownet3d} & $\mathit{Full}$ & 0.114 & 0.412 & 0.771 & 0.602 \\
            HPLFlowNet~\shortcite{gu2019hplflownet} & $\mathit{Full}$& 0.080 & 0.614 &0.855 & 0.429 \\
			PointPWC-Net~\shortcite{wu2019pointpwc} & $\mathit{Full}$ & 0.059 & 0.738 & 0.928 & 0.342 \\
			FLOT~\shortcite{puy20flot} & $\mathit{Full}$ & 0.052 & 0.732 &0.927 & 0.357 \\
			EgoFlow~\shortcite{tishchenko2020self} & $\mathit{Full}$ & 0.069 & 0.670 &0.879 & 0.404 \\ 
			R3DSF~\shortcite{gojcic2021weakly} & $\mathit{Full}$ & 0.052 & 0.746 &  0.936  & 0.361 \\
            PV-RAFT~\shortcite{wei2021pv} & $\mathit{Full}$ & 0.046 & 0.817 &  0.957  &  0.292 \\
            FlowStep3D~\shortcite{Kittenplon_2021_CVPR} & $\mathit{Full}$ & 0.046 & 0.816 &    0.961  &  0.217 \\

		\noalign{\smallskip}\hline\noalign{\smallskip}
			ICP (rigid)~\shortcite{besl1992method} &$\mathit{Self}$& 0.406         & 0.161          & 0.304          & 0.880    \\
            FGR (rigid)~\shortcite{zhou2016fast}        &$\mathit{Self}$& 0.402  & 0.129   & 0.346  & 0.876   \\
        CPD (non-rigid)~\shortcite{myronenko2010point} &$\mathit{Self}$& 0.489  & 0.054   & 0.169  & 0.906   \\
        EgoFlow~\shortcite{tishchenko2020self} & $\mathit{Self}$ & 0.170 & 0.253 &  0.550 &  0.805 \\ 
        PointPWC-Net \shortcite{wu2019pointpwc}   & $\mathit{Self}$& 0.121 & 0.324 & 0.674 & 0.688  \\
		 FlowStep3D~\shortcite{Kittenplon_2021_CVPR} & $\mathit{Self}$ & 0.085 & 0.536 &   0.826  &  \textbf{0.420} \\

    	Ours & $\mathit{Self}$ & \textbf{0.075} & \textbf{0.589} &  \textbf{0.862} & 0.470 \\\hline
    
	\end{tabular}
	
	}
\end{table}

\noindent \textbf{Impact of graph Laplacian regularization:} Encouraging rigid scene estimation via proper regularization is a critical step in handling datasets with many rigid objects. 
\textit{Under-regularization} with an inadequately small $\lambda$ has nearly no impact on original estimation with many possible predictions. 
\textit{Over-regularization} with an excessively large $\lambda$ adds over-strict constraints and leads to imperfect alignment between GMMs.
Results are summarized in Table \ref{tab:laplacian}. Choosing a proper $\lambda$ leads to further improvements on scene flow prediction, resulting in an roughly increase of $5\%$ and $3\%$ in absolute accuracy  on Acc3DS and Acc3DR and about a decrease of  $3\%$ in absolute error on Outliers.

\subsection{Evaluation on FlyingThings3D}\label{sec:ft3d}

We demonstrate its effectiveness by training on the FT3D dataset without using any ground truth annotations.  Table \ref{tab:ft3d} summarizes the evaluation results on the test split of the FT3D dataset. 
Due to the better model design and objective function, our method outperforms all recent self-supervised frameworks including PointPWC-Net \citep{wu2019pointpwc} and FlowStep3D \citep{Kittenplon_2021_CVPR} and some full-supervised methods such as Flownet3D \citep{liu2019flownet3d} and HPLFlowNet \citep{gu2019hplflownet}. 
Specifically, among all self-supervised approaches, our RSFNet obtains the best values on $EPE3D[m]$, $Acc3DS$, and $Acc3DR$,  decreasing  the $EPE3D[m]$ of the previous best model (FlowStep3D) by $11.76\%$ and increasing $Acc3DS$ and $Acc3DR$ of it by $9.88\%$ and $4.36\%$. Note that FlowStep3D has conducted multiple inference steps to iteratively refine the scene flow prediction while our method is more efficient with one-step inference (see the appendix).
\begin{table}[hbt!]

	\centering
	\caption{Evaluation results on the KITTI scene flow datasets.}\label{tab:kitti}
	 \begin{threeparttable}
	\Huge
	\resizebox{\columnwidth}{!}{%
    \begin{tabular}{llcccc}
            \toprule
			Method & Sup. & EPE3D[m]~$\downarrow$ & Acc3DS~$\uparrow$ & Acc3DR~$\uparrow$ & Outliers~$\downarrow$ \\
			\noalign{\smallskip}\hline\noalign{\smallskip}
			Flownet3D~\shortcite{liu2019flownet3d} & $\mathit{Full}$  & 0.177 & 0.374 & 0.668 & 0.527 \\
            HPLFlowNet~\shortcite{gu2019hplflownet} & $\mathit{Full}$   & 0.117 & 0.478 & 0.778 & 0.410 \\
			PointPWC-Net~\shortcite{wu2019pointpwc} & $\mathit{Full}$   & 0.069 & 0.728 & 0.888 & 0.265 \\
			FLOT~\shortcite{puy20flot} & $\mathit{Full}$  & 0.056 & 0.755 & 0.908 & 0.242 \\
			EgoFlow~\shortcite{tishchenko2020self} & $\mathit{Full}$ & 0.103 & 0.488 &0.822 & 0.394 \\
			R3DSF~\shortcite{gojcic2021weakly} & $\mathit{Full}$ & 0.042& 0.849 & 0.959  & 0.208 \\
			PV-RAFT~\shortcite{wei2021pv} & $\mathit{Full}$ & 0.056 &  0.823 &  0.937  &  0.216 \\
			FlowStep3D~\shortcite{Kittenplon_2021_CVPR} & $\mathit{Full}$ & 0.055 & 0.805 &    0.925 &  0.149 \\

		\noalign{\smallskip}\hline\noalign{\smallskip}
			ICP(rigid)~\shortcite{besl1992method}        &$\mathit{Self}$& 0.518         & 0.067          & 0.167          & 0.871          \\
        FGR(rigid)~\shortcite{zhou2016fast}        &$\mathit{Self}$& 0.484  & 0.133   & 0.285  & 0.776  \\
        CPD (non-rigid)~\shortcite{myronenko2010point}        &$\mathit{Self}$& 0.414  & 0.206    & 0.400 & 0.715 \\
        EgoFlow~\shortcite{tishchenko2020self} & $\mathit{Self}$ & 0.415 & 0.221 &   0.372 &   0.810 \\ 
        PointPWC-Net~\shortcite{wu2019pointpwc} &$\mathit{Self}$& \textit{0.255} & \textit{0.238} & \textit{0.496} & \textit{0.686} \\
        FlowStep3D~\shortcite{Kittenplon_2021_CVPR} & $\mathit{Self}$ & 0.102 & 0.708 &   0.839 &  0.246 \\

        Ours & $\mathit{Self}$ & \textbf{0.092} & \textbf{0.747} &  \textbf{0.870} & 0.283 \\

        Ours & $\mathit{Self^*}$ & \textbf{0.096}  & 0.686  & \textbf{0.855}  & 0.302  \\ \hline
	\end{tabular}
	}
	\begin{tablenotes}
	\footnotesize
     \item  \noindent * denotes models trained with KITTI$_r$
   \end{tablenotes}
	 \end{threeparttable}
\end{table}
\subsection{Evaluation on KITTI Scene Flow}

We apply the trained model from the FT3D dataset to the unseen KSF dataset.  As shown in Table \ref{tab:kitti}, our method shows a good generalization result --- achieving a lower 3D end-point error $0.092$ and higher accuracy (\textit{near 4\% absolute accuracy improvement over \citep{Kittenplon_2021_CVPR}}). Models from the KITTI$_r$ also perform competitively though trained on a much smaller dataset containing $6068$ training samples, in contrast to $15,712$ training samples in the FT3D, which shows a promising direction to close the synthetic-to-real gap with real-world training data. More results and details can be found in the appendix.

\section{Conclusions}

In this work, we have presented a novel self-supervised scene flow estimation approach that represents discrete point clouds  as  continuous Gaussian mixture PDFs and recovers motion from their alignment. Models trained with CS show more robustness and higher accuracy compared to CD and EMD. The resulting models have achieved state-of-the-art performance on standard datasets including FT3D and KSF. We are encouraged by this to attempt the design of advanced iterative refinement techniques for better scene flow estimation in future work.

\section{Acknowledgements}

This work is supported by NSF CNS 1922782. The opinions, findings and conclusions expressed in this publication are those of the author(s)
and not necessarily those of the National Science Foundation.

\bibliography{ref}
\clearpage
\newpage

\section{Appendix}
 
\section{Implementation Details}
\subsection{Product of two univariate  Gaussian PDFs}
Given the Gaussian PDF
$\mathcal{N}(x| \mu, \sigma^2) = \frac{1}{\sqrt{2\pi \sigma^2}} e^{-\frac{(x-\mu)^2}{2\sigma^2}}$, the product of two univariate Gaussian PDFs is
\begin{equation}\label{eq:product}
\begin{split}
    \mathcal{N}(x| \mu_1, \sigma_1^2)  \mathcal{N}(x| \mu_2,  \sigma_2^2) = \frac{1}{2\pi \sigma_1 \sigma_2} e^{-\bigg(\frac{(x-\mu_1)^2}{2\sigma_1^2} + \frac{(x-\mu_2)^2}{2\sigma_2^2}\bigg)}
\end{split}
\end{equation}

Let's examine the term in the exponent
\begin{equation}
\begin{split}
    \zeta & = \frac{(x-\mu_1)^2}{2\sigma_1^2} + \frac{(x-\mu_2)^2}{2\sigma_2^2} \\
    & = \frac{(\sigma_1^2+\sigma_2^2)x^2 - 2(\mu_1\sigma_2^2+\mu_2\sigma_1^2)x + \mu_1^2\sigma_2^2+\mu_2^2\sigma_1^2}{2\sigma_1^2\sigma_2^2} \\
    & = \frac{x^2 - 2\frac{\mu_1\sigma_2^2+\mu_2\sigma_1^2}{\sigma_1^2+\sigma_2^2}x + \frac{\mu_1^2\sigma_2^2+\mu_2^2\sigma_1^2}{\sigma_1^2+\sigma_2^2}}{2\frac{\sigma_1^2\sigma_2^2}{\sigma_1^2+\sigma_2^2}} \\
\end{split}
\end{equation}

Let's define
\begin{equation}
    \mu_{12} =  \frac{\mu_1\sigma_2^{2} + \mu_2\sigma_1^{2} }{\sigma_1^{2} + \sigma_2^{2}}
\end{equation}
and
\begin{equation}\label{eq:sigma}
    \sigma_{12}^2 = \frac{\sigma_1^2 \sigma_2^2}{\sigma_1^2 + \sigma_2^2},
\end{equation} we can then rewrite $\zeta$ as:
\begin{equation}
\begin{split}
    \zeta & = \frac{x^2 - 2\frac{\mu_1\sigma_2^2+\mu_2\sigma_1^2}{\sigma_1^2+\sigma_2^2}x + \frac{\mu_1^2\sigma_2^2+\mu_2^2\sigma_1^2}{\sigma_1^2+\sigma_2^2}}{2\frac{\sigma_1^2\sigma_2^2}{\sigma_1^2+\sigma_2^2}} \\
    & = \frac{x^2 - 2\mu_{12}x + \frac{\mu_1^2\sigma_2^2+\mu_2^2\sigma_1^2}{\sigma_1^2+\sigma_2^2}}{2\sigma_{12}^2} \\
    & = \frac{x^2 - 2\mu_{12}x + \mu_{12}^2 +  \bigg(\frac{\mu_1^2\sigma_2^2+\mu_2^2\sigma_1^2}{\sigma_1^2+\sigma_2^2} - \mu_{12}^2\bigg)}{2\sigma_{12}^2} \\
    & = \frac{(x-\mu_{12})^2 }{2\sigma_{12}^2}  + \frac{ \bigg(\frac{\mu_1^2\sigma_2^2+\mu_2^2\sigma_1^2}{\sigma_1^2+\sigma_2^2} - \mu_{12}^2\bigg)}{2\sigma_{12}^2}\\
    & = \frac{(x-\mu_{12})^2 }{2\sigma_{12}^2}  + \frac{ \bigg(\frac{\mu_1^2\sigma_2^2+\mu_2^2\sigma_1^2}{\sigma_1^2+\sigma_2^2} - \mu_{12}^2\bigg)}{2\sigma_{12}^2}\\
    & = \frac{(x-\mu_{12})^2 }{2\sigma_{12}^2}  + \frac{ (\mu_{1} - \mu_{2})^2}{2(\sigma_1^2+\sigma_2^2)}
\end{split}
\end{equation}
Substituting back into Equation \ref{eq:product} gives
\begin{equation}
\begin{split}
\mathcal{N}(x| \mu_1, \sigma_1^2)  \mathcal{N}(x| \mu_2,  \sigma_2^2) & = \frac{1}{2\pi \sigma_1 \sigma_2} e^{-\bigg(\frac{(x-\mu_1)^2}{2\sigma_1^2} + \frac{(x-\mu_2)^2}{2\sigma_2^2}\bigg)} \\
& = \frac{1}{2\pi \sigma_1 \sigma_2}  e^{-\frac{(x-\mu_{12})^2 }{2\sigma_{12}^2}} e^{- \frac{ (\mu_{1} - \mu_{2})^2}{2(\sigma_1^2+\sigma_2^2)}}\label{eq:product_final}
\end{split}
\end{equation}
Note that $\sigma_1 \sigma_2 = \sqrt{\sigma_{12}^2  * (\sigma_1^2 + \sigma_2^2)}$ based on Equation \ref{eq:sigma}. Therefore, we can further write Equation \ref{eq:product_final} as
\begin{equation}
\begin{split}
      \frac{1}{\sqrt{2\pi (\sigma_1^2 + \sigma_2^2})}  e^{- \frac{ (\mu_{1} - \mu_{2})^2}{2(\sigma_1^2+\sigma_2^2)}} \frac{1}{\sqrt{2\pi \sigma_{12}^2}}  e^{-\frac{(x-\mu_{12})^2 }{2\sigma_{12}^2}}
\end{split}
\end{equation}

We now obtain 
\begin{equation}
\footnotesize
\begin{split}\label{eq:single_product}
    \mathcal{N}(x| \mu_1, \sigma_1^2) & \mathcal{N}(x| \mu_2,  \sigma_2^2) = \\ & \mathcal{N}\bigg(\mu_1; \mu_2, \sigma_1^2 + \sigma_2^2 \bigg) \mathcal{N}(x| \mu_{12}, \sigma_{12}^2)
\end{split}
\end{equation}

 \subsection*{Product of two multivariate  Gaussian PDFs}

We now show the derivation of product of two multivariate  Gaussian PDFs, Given two multivariate Gaussians $\mathcal{N}(x|\mu_{1},\Sigma_{1})$
and $\mathcal{N}(x|\mu_{2},\Sigma_{2})$, their product is

\begin{equation}
\begin{split}
\scriptsize
&  \mathcal{N}(x|\mu_{1},\Sigma_{1})\mathcal{N}(x|\mu_{2},\Sigma_{2}) \propto  \\ 
& \exp\left\{ -\frac{1}{2}\left[(x-\mu_{1})^{T}\Sigma_{1}^{-1}(x-\mu_{1}) +  (x-\mu_{2})^{T}\Sigma_{2}^{-1}(x-\mu_{2})\right]\right\}.
\end{split}
\end{equation}
The exponent is then a sum of two quadratic forms. This can be simplified.
\begin{equation}
\begin{split}
\footnotesize
(x- &  \mu_{1})^{T} \Sigma_{1}^{-1}(x-\mu_{1})+(x-\mu_{2})^{T}\Sigma_{2}^{-1}(x-\mu_{2}) \\
& =  x^{T}\Sigma_{1}^{-1}x+x^{T}\Sigma_{2}^{-1}x+\mu_{1}^{T}\Sigma_{1}^{-1}\mu_{1}+\mu_{2}^{T}\Sigma_{2}^{-1}\mu_{2}  \\ & \quad -2\mu_{1}^{T}\Sigma_{1}^{-1}x-2\mu_{2}^{T}\Sigma_{2}^{-1}x \\
& =x^{T}\left(\Sigma_{1}^{-1}+\Sigma_{2}^{-1}\right)x-2\left(\Sigma_{1}^{-1}\mu_{1}+\Sigma_{2}^{-1}\mu_{2}\right)^{T}x\\ & \quad +\mu_{1}^{T}\Sigma_{1}^{-1}\mu_{1}+\mu_{2}^{T}\Sigma_{2}^{-1}\mu_{2}.
\end{split}
\end{equation}
Let's define $\mu=\left(\Sigma_{1}^{-1}+\Sigma_{2}^{-1}\right)^{-1}\left(\Sigma_{1}^{-1}\mu_{1}+\Sigma_{2}^{-1}\mu_{2}\right)$ and complete squares, we then get
\begin{equation}
\begin{split}
\footnotesize
&(x-\mu_{1})^{T}\Sigma_{1}^{-1}(x-\mu_{1})+(x-\mu_{2})^{T}\Sigma_{2}^{-1}(x-\mu_{2}) \\
& =\left(x-\mu\right)^{T}\left(\Sigma_{1}^{-1}+\Sigma_{2}^{-1}\right)^{-1}\left(x-\mu\right) \\ & \quad -C+\mu_{1}^{T}\Sigma_{1}^{-1}\mu_{1}+\mu_{2}^{T}\Sigma_{2}^{-1}\mu_{2}, \label{eq:squares}
\end{split}
\end{equation}
where 
\begin{equation}
\footnotesize
C\equiv\left(\Sigma_{1}^{-1}\mu_{1}+\Sigma_{2}^{-1}\mu_{2}\right)^{T}\left(\Sigma_{1}^{-1}+\Sigma_{2}^{-1}\right)^{-1}\left(\Sigma_{1}^{-1}\mu_{1}+\Sigma_{2}^{-1}\mu_{2}\right).
\end{equation}
Expanding $C$, we get 
\begin{equation}
\begin{split}
\footnotesize
C= & \mu_{1}^{T}\Sigma_{1}^{-1}\left(\Sigma_{1}^{-1}+\Sigma_{2}^{-1}\right)^{-1}\Sigma_{1}^{-1}\mu_{1} \\ + &\mu_{2}^{T}\Sigma_{2}^{-1}\left(\Sigma_{1}^{-1}+\Sigma_{2}^{-1}\right)^{-1}\Sigma_{2}^{-1}\mu_{2}\\ + & 2\mu_{2}^{T}\Sigma_{2}^{-1}\left(\Sigma_{1}^{-1}+\Sigma_{2}^{-1}\right)^{-1}\Sigma_{1}^{-1}\mu_{1}.
\end{split}
\end{equation}
Using the Woodbury formula $(A^{-1}+B^{-1})^{-1}=A-A(A+B)^{-1}A$, i.e., let $A = \Sigma_{1}^{-1}$ and $B = \Sigma_{2}^{-1}$, we get
\begin{equation}
\begin{split}
\footnotesize
C & = \mu_{1}^{T}\left[\Sigma_{1}^{-1}-\left(\Sigma_{1}+\Sigma_{2}\right)^{-1}\right]\mu_{1} \\
 & +\mu_{2}^{T}\left[\Sigma_{2}^{-1}-\left(\Sigma_{1}+\Sigma_{2}\right)^{-1}\right]\mu_{2}\\
 &+2\mu_{2}^{T}(\Sigma_{1}+\Sigma_{2})^{-1}\mu_{1}.
\end{split}
\end{equation}
Substituting this in (\ref{eq:squares}), we get
\begin{equation}
\begin{split}
\footnotesize
& \xi = (x-\mu_{1})^{T}\Sigma_{1}^{-1}(x-\mu_{1})+(x-\mu_{2})^{T}\Sigma_{2}^{-1}(x-\mu_{2}) \\
& =\left(x-\mu\right)^{T}\left(\Sigma_{1}^{-1}+\Sigma_{2}^{-1}\right)^{-1}\left(x-\mu\right) \\ & \quad + \left(\mu_{1}-\mu_{2}\right)^{T}\left(\Sigma_{1}+\Sigma_{2}\right)^{-1}\left(\mu_{1}-\mu_{2}\right).
\end{split}
\end{equation}
Therefore the product of Gaussians can be written as

\begin{equation}
\begin{split}
\footnotesize
& \mathcal{N}(x|\mu_{1},\Sigma_{1})\mathcal{N}(x|\mu_{2},\Sigma_{2}) \\  & =\frac{1}{\sqrt{\left|(2\pi)\Sigma_{1}\right|\left|(2\pi)\Sigma_{2}\right|}}\exp\left\{ -\frac{1}{2}\xi \right\} \\ &
=\frac{\sqrt{\left|\left(\Sigma_{1}^{-1}+\Sigma_{2}^{-1}\right)\right|}}{\sqrt{\left|(2\pi)\Sigma_{1}\right|\left|(2\pi)\Sigma_{2}\right|\left|\left(\Sigma_{1}^{-1}+\Sigma_{2}^{-1}\right)\right|}}\exp\left\{ -\frac{1}{2}\xi \right\} .
\end{split}
\end{equation}
We now simplify the denominator. Since $\left|AB\right|=\left|BA\right|=\left|A\right|\left|B\right|$
we have $\left|\Sigma_{1}\right|\left|\Sigma_{2}\right|\left|\left(\Sigma_{1}^{-1}+\Sigma_{2}^{-1}\right)\right|=\left|\Sigma_{1}\left(\Sigma_{1}^{-1}+\Sigma_{2}^{-1}\right)\Sigma_{2}\right|=\left|\left(\Sigma_{1}+\Sigma_{2}\right)\right|$.
Therefore, after defining $\Sigma\equiv\left(\Sigma_{1}^{-1}+\Sigma_{2}^{-1}\right)^{-1}$,
we have 
\begin{equation}
\begin{split}
\footnotesize
&  \mathcal{N}(x|\mu_{1},\Sigma_{1})\mathcal{N}(x|\mu_{2},\Sigma_{2}) \\
& =  \frac{1}{\sqrt{\left|(2\pi)\left(\Sigma_{1}+\Sigma_{2}\right)\right|\left|2\pi\Sigma^{-1}\right|}}\exp\left\{ -\frac{1}{2}\xi \right\} 
\\ & =\mathcal{N}(x|\mu,\Sigma)\mathcal{N}(\mu_{1}|\mu_{2},\Sigma_{1}+\Sigma_{2}). \label{eq:multivariate_product}
\end{split}
\end{equation}

Equation \ref{eq:main_product} in the main paper is the multivariate version of Equation \ref{eq:multivariate_product}. We then obtain the product of two multivariate Gaussian PDFs as
\begin{equation}
\begin{split}\label{eq:product_appendix}
    \mathcal{N}(\boldsymbol{x}| \boldsymbol{\mu_i}, \boldsymbol{\Sigma_i}) & \mathcal{N}(\boldsymbol{x}| \boldsymbol{\mu_j}, \boldsymbol{\Gamma_j})  = \\ & \mathcal{N}(\boldsymbol{\mu_i}| \boldsymbol{\mu_j}, \boldsymbol{\Sigma_i}  + \boldsymbol{\Gamma_j}) \mathcal{N}(x| \boldsymbol{\mu_{ij}}, \boldsymbol{\Sigma_{ij}})
\end{split}
\end{equation}
 where
 \begin{equation}
 \footnotesize
     \boldsymbol{\mu_{ij}}  = \boldsymbol{\Sigma_{ij}} (\boldsymbol{\Sigma_i}^{-1} \boldsymbol{\mu_{i}}  + \boldsymbol{\Gamma_j}^{-1} \boldsymbol{\mu_{j}})
 \end{equation}
and
\begin{equation}
 \footnotesize
\boldsymbol{\Sigma_{ij}} = {(\boldsymbol{\Sigma_i}^{-1}  + \boldsymbol{\Gamma_j}^{-1})}^{-1}.
\end{equation} 

\subsection{Closed-form Expression for the CS divergence}

Inspired by the CS inequality, the CS divergence measure \cite{jenssen2005optimizing} is defined as 
\begin{equation}
\footnotesize
\begin{split}
    D_{CS}(q,p) & = - \log \Big( \frac{\int q(x) p(x) dx}{\sqrt{\int q(x)^2 dx \int p(x)^2 dx}} \Big)  \\
  & = - \log  \int  q(x) p(x) dx +  0.5 \log  \int q(x)^2 dx \\ &  \quad   +  0.5 \log \int p(x)^2 dx
\end{split}
\end{equation}
It defines a symmetric measure for any two PDFs $q$ and $p$ such that $0 \le D_{CS} < \infty$ where the mimimum is obtained iff $q(x) = p(x)$.

The CS divergence can be written in a closed-form expression for GMMs \cite{jenssen2006cauchy}. Formally, for a given point cloud $\boldsymbol{x}$, we represent it as the PDF of a general Gaussian mixture, which is defined as  $\mathcal{G}(x) = \sum_{k=1}^K w_k \mathcal{N}(x | \boldsymbol{\mu_k}, \boldsymbol{\Sigma_k})$, where
\begin{equation}
    \mathcal{N}(x| \boldsymbol{\mu_k}, \boldsymbol{\Sigma_k}) = \frac{\exp\left[-\frac{1}{2}(x-\boldsymbol{\mu_k})^T \boldsymbol{\Sigma_k}^{-1}(x-\boldsymbol{\mu_k})\right]}{\sqrt{ (2\pi)^d |\boldsymbol{\Sigma_k}|}}
\end{equation}
where $K$ is the number of Gaussian components. We denote $w_k, \mu_{k}, \Sigma_{k}$  as the mixture coefficient, mean, and covariance matrix  of the $k^{th}$ component of $\mathcal{G}(x)$. 
$d$ is the feature dimension of each point. 
In our case, $d=3$. $|\boldsymbol{\Sigma_k}|\equiv \det \boldsymbol{\Sigma_k}$ is the determinant of $\boldsymbol{\Sigma_k}$, also known as the generalized variance. 

Given the warped point cloud $\boldsymbol{S}_{w} =\boldsymbol{S}+\boldsymbol{D}$ and the target point cloud $\boldsymbol{T}$, we represent $\boldsymbol{S}_{w}$ and $\boldsymbol{T}$ as the GMM representations $\mathcal{G}(\boldsymbol{S}_{w})$ and $\mathcal{G}(\boldsymbol{T})$:
\begin{align}
    \mathcal{G}(\boldsymbol{S}_{w}) = \sum_{i=1}^N \pi_{i}  \mathcal{N}(x | \boldsymbol{c^s_i}, \boldsymbol{\Sigma_i})
\end{align}
and
\begin{align}
\footnotesize
    \mathcal{G}(\boldsymbol{T}) = \sum_{j=1}^M \tau_{j} \mathcal{N}(x|  \boldsymbol{c^t_j}, \Gamma_j)
\end{align} where we denote the sets of mixture coefficients for two GMMs $\mathcal{G}(\boldsymbol{S}_{w})$ and $\mathcal{G}(\boldsymbol{T})$ as $\{ \pi_{i} \}_{i=1}^N$ and $\{ \tau_{j} \}_{j=1}^M$ and their covariance matrix sets as $\{ \Sigma_{i} \}_{i=1}^N$ and $\{ \Gamma_{i} \}_{j=1}^M$. 

The CS divergence between $\mathcal{G}(\boldsymbol{S}_{w})$ and $\mathcal{G}(\boldsymbol{T})$ is defined as
\begin{equation}
\footnotesize
\begin{split}\label{eq:cs_divergence_appendix}
    & \mathcal{D}_{CS}( \mathcal{G}(\boldsymbol{S}_{w}),\mathcal{G}(\boldsymbol{T}))  = - \log \Big( \frac{\int \mathcal{G}(\boldsymbol{S}_{w}) \mathcal{G}(\boldsymbol{T}) dx}{\sqrt{\int {\mathcal{G}(\boldsymbol{S}_{w})}^2 dx \int {\mathcal{G}(\boldsymbol{T})}^2 dx}} \Big)  \\
   & = - \log  \int  \mathcal{G}(\boldsymbol{S}_{w})   \mathcal{G}(\boldsymbol{T}) dx +  0.5 \log  \int {\mathcal{G}(\boldsymbol{S}_{w})}^2 dx \\ &  \quad +  0.5 \log   \int  {\mathcal{G}(\boldsymbol{T})}^2 dx
\end{split}
\end{equation}

Using the Gaussian identity in Equation \ref{eq:product_appendix}, we can write a closed-form
expression of $\log \int  \mathcal{G}(\boldsymbol{S}_{w})   \mathcal{G}(\boldsymbol{T}) dx $ as follows:
\begin{equation}
\footnotesize
\begin{split}
    & \log \bigg(\int \sum_{i=1}^N \sum_{j=1}^M \pi_{i}  \tau_{j} \mathcal{N}(x | \boldsymbol{c^s_i}, \boldsymbol{\Sigma_i}) \mathcal{N}(x|  \boldsymbol{c^t_j}, \Gamma_j) dx\bigg) \\
    =& \log \bigg(\sum_{i=1}^N \sum_{j=1}^M \pi_{i}  \tau_{j} \int \mathcal{N}(x | \boldsymbol{c^s_i}, \boldsymbol{\Sigma_i}) \mathcal{N}(x|  \boldsymbol{c^t_j}, \Gamma_j) dx\bigg)\\
    =& \log \bigg(\sum_{i=1}^N \sum_{j=1}^M \pi_{i}  \tau_{j} \int \mathcal{N}(\boldsymbol{c^s_i}| \boldsymbol{c^t_j}, \boldsymbol{\Sigma_i}  + \boldsymbol{\Gamma_j}) \mathcal{N}(x| \boldsymbol{\mu_{ij}}, \boldsymbol{\Sigma_{ij}}) dx\bigg)\\
    =& \log \bigg(\sum_{i=1}^N \sum_{j=1}^M \pi_{i}  \tau_{j}  \mathcal{N}(\boldsymbol{c^s_i}| \boldsymbol{c^t_j}, \boldsymbol{\Sigma_i}  + \boldsymbol{\Gamma_j}) \underbrace{ \int \mathcal{N}(x| \boldsymbol{\mu_{ij}}, \boldsymbol{\Sigma_{ij}}) dx}_{=1} \bigg)\\
    =& \log \bigg(\sum_{i=1}^N \sum_{j=1}^M \pi_{i}  \tau_{j}  \mathcal{N}(\boldsymbol{c^s_i}| \boldsymbol{c^t_j}, \boldsymbol{\Sigma_i}  + \boldsymbol{\Gamma_j}) \bigg) 
\end{split}
\end{equation}

Applying the same trick to the second and third term of  Equation \ref{eq:cs_divergence_appendix}, we get
\begin{equation}
\footnotesize
\begin{split}
 & \mathcal{D}_{CS}( \mathcal{G}(\boldsymbol{S}_{w}),\mathcal{G}(\boldsymbol{T})) =  - \log \bigg(\sum_{i=1}^N \sum_{j=1}^M \pi_{i}  \tau_{j}  \mathcal{N}(\boldsymbol{c^s_i}| \boldsymbol{c^t_j}, \boldsymbol{\Sigma_i}  + \boldsymbol{\Gamma_j}) \bigg)  \\
         & + 0.5 \log \bigg(\sum_{i=1}^N \sum_{i'=1}^N \pi_{i}  \pi_{i'} \mathcal{N}(\boldsymbol{c^s_i}| \boldsymbol{c^s_{i'}}, \boldsymbol{\Sigma_i}  + \boldsymbol{\Sigma_{i'}}) \bigg) \\ 
         & + 0.5 \log \bigg(\sum_{j=1}^M \sum_{j'=1}^M \tau_{j}  \tau_{j'}  \mathcal{N} (\boldsymbol{c^t_j} | \boldsymbol{c^t_{j'}}, \boldsymbol{\Gamma_j} + \boldsymbol{\Gamma_{j'}}) \bigg).
\end{split}
\end{equation}

\subsection{Implementation of the CS Divergence Loss}
The CS divergence loss can be implemented with a few lines of code. To handle the numerical issue, we leverage the Log-Sum-Exp trick as shown in Algorithm \ref{alg:code}. 
\begin{algorithm}[hbt!]
\caption{The CS divergence implemented in PyTorch.}
\label{alg:code}
\definecolor{codeblue}{rgb}{0.25,0.5,0.5}
\lstset{
  backgroundcolor=\color{white},
  basicstyle=\fontsize{7.2pt}{7.2pt}\ttfamily\selectfont,
  columns=fullflexible,
  breaklines=true,
  captionpos=b,
  commentstyle=\fontsize{7.2pt}{7.2pt}\color{codeblue},
  keywordstyle=\fontsize{7.2pt}{7.2pt},
}
\begin{lstlisting}[language=python]
# est_flow. Predicted flow <-- B X N x 3
# source, target. point coordinates <-- B X N x 3 and B X M x 3
# sigma, gamma. Isotropic variances <-- scalar
# tau, nu. Mixture coefficients <-- scalar
def GMM(c_s, c_t, sigma, gamma, tau, nu):
    p_ij = tau * nu
    sigma_ij = sigma + gamma
    # B X N x M x 3  <-- B X N X 1 x 3 - B X 1 X M X 3
    diff_ij = (c_s.unsqueeze(2) - c_t.unsqueeze(1))
    # B X N x M
    diff_ij = (diff_ij**2).sum(-1).div(sigma_ij).mul(-0.5) - 1.5*log(2*math.pi) - 1.5*math.log(sigma_ij) + math.log(p_ij)
    # the log_sum_exp trick
    dist = torch.logsumexp((diff_ij).reshape(diff_ij.shape[0], -1),dim=1).mean()
    return dist

def cs_divergence_loss(source, target, est_flow, sigma, gamma, tau, nu):
    c_s = source + est_flow
    c_t = target
    st_dist = -1 * GMM(c_s, c_t, sigma, gamma, tau, nu)
    ss_dist = 0.5 * GMM(c_s, c_s, sigma, sigma, tau, tau)
    tt_dist = 0.5 * GMM(c_t, c_t, gamma, gamma, nu, nu)
    cs_divergence = ss_dist + st_dist + tt_dist
    return cs_divergence
\end{lstlisting}
\end{algorithm}

\begin{figure*}[hbt!]
    \centering
     \includegraphics[width=\textwidth]{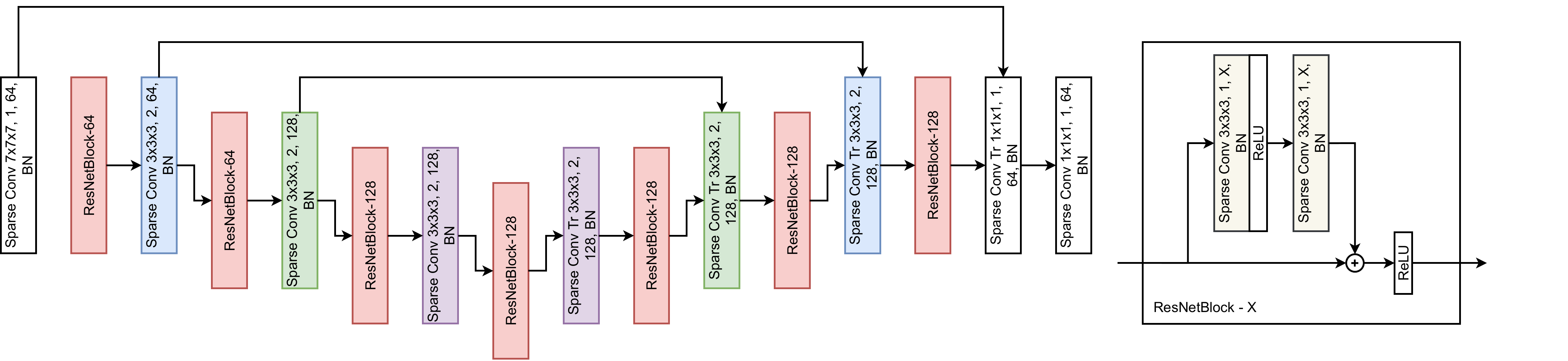}
    \caption{Network architecture of the scene flow network.  Both sparse convolutional (Sparse Conv) and sparse transpose convolutational (Sparse Conv Tr) layers are applied, where 3D kernel size, stride, output feature dimension, and normalization functions are denoted, e.g., 3x3x3, 2, 64, BN. BN is the batch normalization. (best viewed on display)}
    \label{fig:architecture}
\end{figure*}

\subsection{Run Time}

We compare our method against state-of-the-art self-supervised models including PointPWC-Net \citep{wu2019pointpwc} and FlowStep3D \citep{Kittenplon_2021_CVPR}. We use the official implementations released by the authors and evaluate all models on a server equipped with AMD EPYC ROME and NVIDIA A100 GPUs. PointPWCNet \citep{wu2019pointpwc} contains roughly 7.7 million parameters while requiring $0.058$ seconds on average for one inference step. FlowStep3D \citep{Kittenplon_2021_CVPR} has a lowest number of model parameters ($0.689$ million). However, FlowStep3D \citep{Kittenplon_2021_CVPR} takes a longer time ($0.572$ seconds) to process one point cloud pair due to its multiple inference iterations. Our RSFNet has approximately 7.8 million parameters and takes $0.083$ seconds on average for processing one point cloud pair while generating accurate and robust predictions.

\subsection{Network Architectures}

Our model is built upon Pytorch \citep{paszke2019pytorch} and MinkowskiEngine \cite{choy20194d}. The backbone architecture is depicted in Figure \ref{fig:architecture}. We feed the absolute point coordinates of the point cloud pairs into the model. All the sparse convolution layers are followed by a \textit{batch normalization} layer and a \textit{ReLU} activation function in ResNet blocks. We obtain the output of 64-dimensional pointwise latent features, which are further passed to the scene flow estimation module.

\section{Additional Evaluation Results}

In this section, we provide additional evaluation results on FT3D and KSF that were omitted from the main paper due to the space constraint. 

\subsection{Additional Results on FT3D}
We evaluate the fully supervised performance of the designed backbone architecture trained with ground truth annotations on FT3D, which supplements the evaluation of Table \ref{tab:ft3d} in the Experiment Section. The results are summarized in Table \ref{tab:ft3d_additional}. As expected, our model achieves competitive results compared to existing fully supervised approaches. 
\begin{table}[hbt!]
	\centering
	\caption{Additional evaluation results on \emph{FT3D}. CS achieves a better self-supervised performance over CD and EMD.}\label{tab:ft3d_additional}
	\Huge
	\resizebox{\columnwidth}{!}{%
    \begin{tabular}{lccccc}
            \toprule
			Method & Sup. & EPE3D[m]~$\downarrow$ & Acc3DS~$\uparrow$ & Acc3DR~$\uparrow$ & Outliers~$\downarrow$ \\
			\noalign{\smallskip}\hline\noalign{\smallskip}
             FlowNet3D~\shortcite{liu2019flownet3d} & $\mathit{Full}$ & 0.114 & 0.412 & 0.771 & 0.602 \\
            HPLFlowNet~\shortcite{gu2019hplflownet} & $\mathit{Full}$& 0.080 & 0.614 &0.855 & 0.429 \\
			PointPWC-Net~\shortcite{wu2019pointpwc} & $\mathit{Full}$ & 0.059 & 0.738 & 0.928 & 0.342 \\
			FLOT~\shortcite{puy20flot} & $\mathit{Full}$ & 0.052 & 0.732 &0.927 & 0.357 \\
			EgoFlow~\shortcite{tishchenko2020self} & $\mathit{Full}$ & 0.069 & 0.670 &0.879 & 0.404 \\ 
			R3DSF~\shortcite{gojcic2021weakly} & $\mathit{Full}$ & 0.052 & 0.746 &  0.936  & 0.361 \\
            PV-RAFT~\shortcite{wei2021pv} & $\mathit{Full}$ & 0.046 & 0.817 &  0.957  &  0.292 \\
            FlowStep3D~\shortcite{Kittenplon_2021_CVPR} & $\mathit{Full}$ & 0.046 & 0.816 &    0.961  &  0.217 \\
             \noalign{\smallskip}\hline\noalign{\smallskip}
			Ours & $\mathit{Full}$ & 0.052 & 0.746 &  0.932  & 0.361 \\ \hline
            \noalign{\smallskip}\hline\noalign{\smallskip}
			ICP (rigid)~\shortcite{besl1992method} &$\mathit{Self}$& 0.406         & 0.161          & 0.304          & 0.880    \\
            FGR (rigid)~\shortcite{zhou2016fast}        &$\mathit{Self}$& 0.402  & 0.129   & 0.346  & 0.876   \\
        CPD (non-rigid)~\shortcite{myronenko2010point} &$\mathit{Self}$& 0.489  & 0.054   & 0.169  & 0.906   \\
        EgoFlow~\shortcite{tishchenko2020self} & $\mathit{Self}$ & 0.170 & 0.253 &  0.550 &  0.805 \\ 
        PointPWC-Net \shortcite{wu2019pointpwc}   & $\mathit{Self}$& 0.121 & 0.324 & 0.674 & 0.688  \\
		 FlowStep3D~\shortcite{Kittenplon_2021_CVPR} & $\mathit{Self}$ & 0.085 & 0.536 &   0.826  &  0.420 \\
             \noalign{\smallskip}\hline\noalign{\smallskip}
         Ours (CD)   & $\mathit{Self}$ & 0.112  & 0.347 & 0.665 & 0.632 \\
		  Ours (EMD)   & $\mathit{Self}$ & 0.121 & 0.332 & 0.617 & 0.637 \\
		 
		 Ours (CS)   & $\mathit{Self}$ & \textbf{0.109} & \textbf{0.365} &  \textbf{0.671} & \textbf{0.612} \\ \hline

	\end{tabular}
	
	}
\end{table}
\begin{table}
	\centering
	\caption{Additional evaluation results on KSF.}
	\Huge
	\resizebox{\columnwidth}{!}{%
    \begin{tabular}{llcccc}
            \toprule
			Method & Sup. & EPE3D[m]~$\downarrow$ & Acc3DS~$\uparrow$ & Acc3DR~$\uparrow$ & Outliers~$\downarrow$ \\
			\noalign{\smallskip}\hline\noalign{\smallskip}
			Flownet3D~\shortcite{liu2019flownet3d} & $\mathit{Full}$  & 0.177 & 0.374 & 0.668 & 0.527 \\
            HPLFlowNet~\shortcite{gu2019hplflownet} & $\mathit{Full}$   & 0.117 & 0.478 & 0.778 & 0.410 \\
			PointPWC-Net~\shortcite{wu2019pointpwc} & $\mathit{Full}$   & 0.069 & 0.728 & 0.888 & 0.265 \\
			FLOT~\shortcite{puy20flot} & $\mathit{Full}$  & 0.056 & 0.755 & 0.908 & 0.242 \\
			EgoFlow~\shortcite{tishchenko2020self} & $\mathit{Full}$ & 0.103 & 0.488 &0.822 & 0.394 \\
			R3DSF~\shortcite{gojcic2021weakly} & $\mathit{Full}$ & 0.042& 0.849 & 0.959  & 0.208 \\
			PV-RAFT~\shortcite{wei2021pv} & $\mathit{Full}$ & 0.056 &  0.823 &  0.937  &  0.216 \\
			FlowStep3D~\shortcite{Kittenplon_2021_CVPR} & $\mathit{Full}$ & 0.055 & 0.756 &    0.935 &  0.353 \\
		    \noalign{\smallskip}\hline\noalign{\smallskip}
			Ours & $\mathit{Full}$ & 0.078 & 0.770 &  0.891  & 0.268 \\ \hline
	\end{tabular}
	}
\end{table}

\subsection{Additional Results on KSF}

We further apply the model trained on FT3D with full annotations to the unseen KSF dataset, reflecting its performance in a real-world environment. The model still performs competitively without adding any refinement module adopted in \citep{puy20flot,gojcic2021weakly}.

\begin{table}[H]
	\centering
	\caption{Comparisons between FLOT \citep{puy20flot} models trained with  CD, EMD, and CS. We  evaluate all trained  models on the KSF dataset. CS provides a robust scene flow prediction which is reflected by its lower value on Outliers and higher values on Acc3DS and Acc3DR. The resulting models outperform some representative self-supervised approaches.}\label{tab:robustness_flot}
	\Huge
	\resizebox{\columnwidth}{!}{%
    \begin{tabular}{l|cc|cccc}
            \toprule
			Method & Sup. & Training data &EPE3D [m]~$\downarrow$ & Acc3DS~$\uparrow$ & Acc3DR~$\uparrow$ & Outliers~$\downarrow$ \\
			\noalign{\smallskip}\hline\noalign{\smallskip}
	        ICP(rigid)~\shortcite{besl1992method}        &$\mathit{Self}$ & FT3D & 0.518         & 0.067          & 0.167          & 0.871          \\
        FGR(rigid)~\shortcite{zhou2016fast}        &$\mathit{Self}$ & FT3D & 0.484  & 0.133   & 0.285  & 0.776  \\
        CPD (non-rigid)~\shortcite{myronenko2010point}        &$\mathit{Self}$ & FT3D & 0.414  & 0.206    & 0.400 & 0.715 \\
        EgoFlow~\shortcite{tishchenko2020self} & $\mathit{Self}$& FT3D  & 0.415 & 0.221 &   0.372 &   0.810 \\  
        	\noalign{\smallskip}\hline\noalign{\smallskip}

            FLOT \shortcite{puy20flot} + CD (Ours) & $\mathit{Self}$ & KITTI$_r$& 0.416 & 0.205 &  0.397 &  0.687 \\
            FLOT \shortcite{puy20flot} + EMD (Ours)  & $\mathit{Self}$ & KITTI$_r$  & \textbf{0.358} & 0.282 &  0.484 &  0.616 \\
            FLOT \shortcite{puy20flot} + CS (Ours) & $\mathit{Self}$ & KITTI$_r$ & 0.396 & \textbf{0.325} &   \textbf{0.511} &  \textbf{0.592} \\
            \hline
	\end{tabular}
	}

\end{table}

\begin{figure*}[t]
    \centering
    \includegraphics[width=\textwidth]{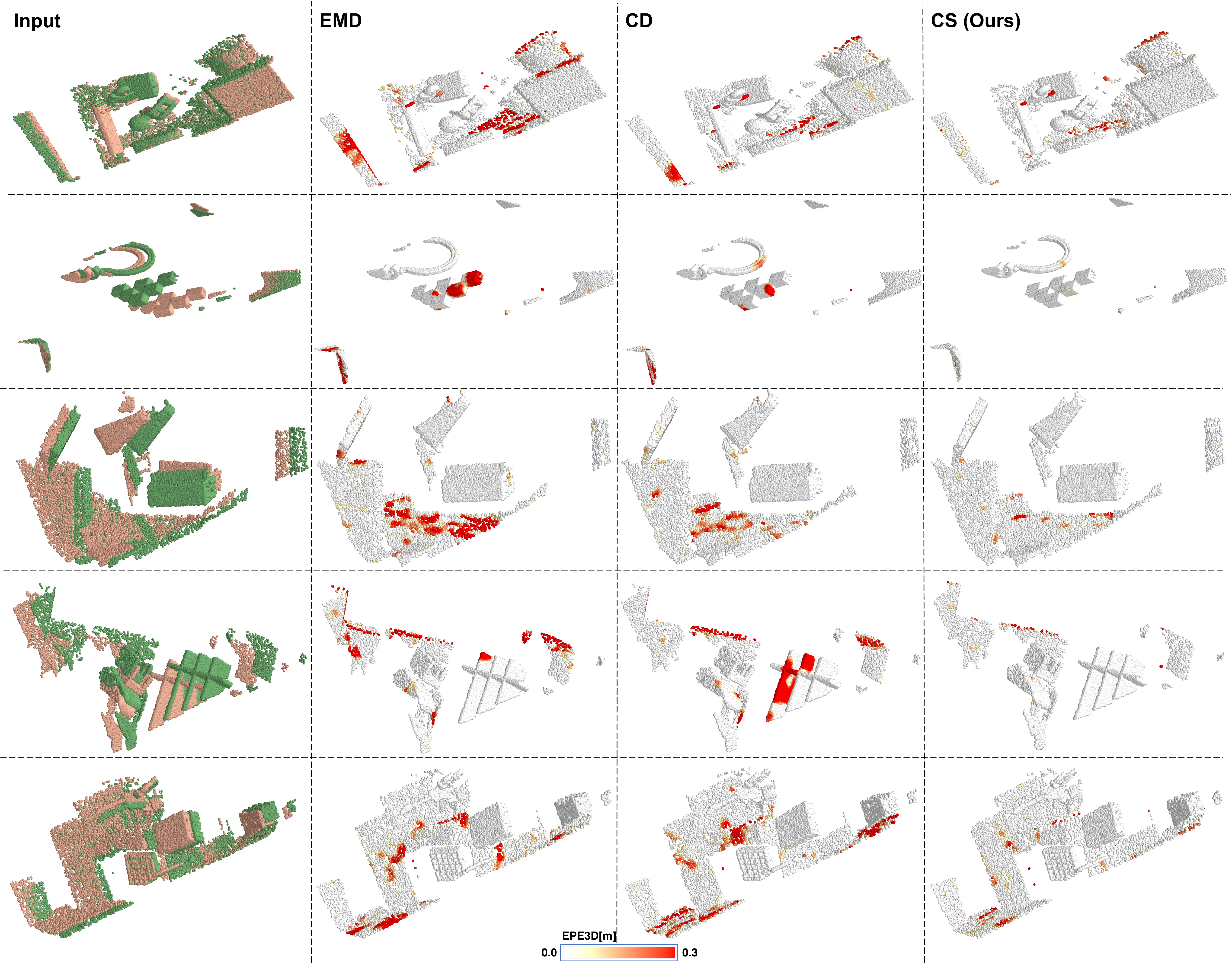}
    \caption{Qualitative results of our method on \emph{FT3D}. We clip the EPE3D[m] to the range between $0.0$ m (white) and $0.3$ (red). It confirms that our method is better on handling outliers compared to EMD and CD. }\label{fig:robustness_additional}
\end{figure*}

\subsection{Comparison with CD and EMD with FT3D}

We train models with CD, EMD, and CS on the FT3D dataset (Table \ref{tab:ft3d_additional}). FT3D is generated in a synthetic environment, containing less noisy points compared to the real-world KITTI$_{r}$. CS consistently outperforms CD and EMD, drawing a similar conclusion of Table \ref{tab:robustness}.  For example, the CD has about $2\%$ absolute performance drop in Acc3DS and Outliers compared to CS. More qualitative examples can be found in Figure \ref{fig:robustness_additional}.
\subsection{Comparison with CD and EMD in FLOT}

We apply CS to the popular FLOT model \citep{puy20flot}, which finds the soft correspondence between points based on optimal transport \citep{peyre2019computational}. We didn't use its refinement module as it hurts the performance when conducting self-supervised learning. The results are summarized in Table \ref{tab:robustness_flot}.

\subsection{Effects of Number of Input Points}
We show the model's performance change w.r.t. the number of input points in Table \ref{tab:robustness_points}. As expected, the performance of all models grows consistently as we increase the number of points. And  CS outperforms CS and EMD under this setting.

\begin{table}[H]
	\centering
	\caption{Model performance with different number of input points. We  evaluate all trained  models on the KSF dataset.}\label{tab:robustness_points}
	\Huge
	\resizebox{\columnwidth}{!}{%
    \begin{tabular}{l|ccc|cccc}
            \toprule
			Method & Sup. & Training data & Sampling &EPE3D [m]~$\downarrow$ & Acc3DS~$\uparrow$ & Acc3DR~$\uparrow$ & Outliers~$\downarrow$ \\
			\noalign{\smallskip}\hline\noalign{\smallskip}
			CD (Ours) & $\mathit{Self}$ & KITTI$_r$ &  1,024 & 0.507       & 0.084         & 0.227         &0.862   \\
            EMD (Ours)   & $\mathit{Self}$ & KITTI$_r$  & 1,024  & 0.586  & 0.078   & 0.224  & 0.864   \\
            CS (Ours) & $\mathit{Self}$ & KITTI$_r$ & 1,024 & \textbf{ 0.369} & \textbf{0.177} &   \textbf{0.395} &  \textbf{0.716} \\
            \noalign{\smallskip}\hline\noalign{\smallskip}
            CD (Ours) & $\mathit{Self}$ & KITTI$_r$ &  2,048 & 0.507       & 0.084         & 0.227         &0.862   \\
            EMD (Ours)   & $\mathit{Self}$ & KITTI$_r$  & 2,048  & 0.299  & 0.231  & 0.486  & 0.658  \\
            CS (Ours) & $\mathit{Self}$ & KITTI$_r$ & 2,048 & \textbf{ 0.184} & \textbf{0.419} &   \textbf{0.673} &  \textbf{0.492} \\
            \noalign{\smallskip}\hline\noalign{\smallskip}
            CD (Ours) & $\mathit{Self}$ & KITTI$_r$ & 4,096  & 0.286 & 0.314 &   0.525 &  0.624 \\
            EMD (Ours)   & $\mathit{Self}$ & KITTI$_r$  & 4,096  & 0.294 & 0.309 &   0.549 & 0.606 \\
            CS (Ours) & $\mathit{Self}$ & KITTI$_r$ & 4,096  & \textbf{0.171} & \textbf{0.480} &   \textbf{0.716} &  \textbf{0.449} \\
            \noalign{\smallskip}\hline\noalign{\smallskip}
            CD (Ours) & $\mathit{Self}$ & KITTI$_r$ & 8,192 & 0.170         & 0.477         & 0.697         & 0.470   \\
            EMD (Ours)   & $\mathit{Self}$ & KITTI$_r$ & 8,192 & 0.192  & 0.426   & 0.666  & 0.503   \\
            CS (Ours) & $\mathit{Self}$ & KITTI$_r$ & 8,192 & \textbf{0.105} & \textbf{0.633} &   \textbf{0.832} &  \textbf{0.338} \\
             \bottomrule
	\end{tabular}
	}

\end{table}

\end{document}